\documentclass[manuscript, nonacm, screen]{acmart}
\usepackage{booktabs} 
\usepackage{tabu}

\usepackage[ruled]{algorithm2e} 

\acmDOI{0000001.0000001}

\begin{document}
\title{Robot Assisted Tower Construction - A Resource Distribution Task to Study Human-Robot Collaboration and Interaction with Groups of People}

\author{Malte F. Jung}
\affiliation{%
  \institution{Cornell University}
  \city{Ithaca}
  \state{NY}
  \postcode{14850}
  \country{USA}}
\email{mjung@cornell.edu}

\author{Dominic DiFranzo}
\affiliation{%
  \institution{Cornell University}
  \city{Ithaca}
  \state{NY}
  \postcode{14850}
  \country{USA}}
\email{djd274@cornell.edu}

\author{Brett Stoll}
\affiliation{%
  \institution{Cornell University}
  \city{Ithaca}
  \state{NY}
  \postcode{14850}
  \country{USA}}
\email{bas364@cornell.edu}

\author{Solace Shen}
\affiliation{%
  \institution{Robinhood}
  \city{Menlo Park}
  \state{CA}
  \postcode{94025}
  \country{USA}}
\email{solaceshen@gmail.com}

\author{Austin Lawrence}
\affiliation{%
  \institution{RoBotany}
  \city{Pittsburgh}
  \state{PA}
  \postcode{15213}
  \country{USA}}
\email{ablarry91@gmail.com}

\author{Houston Claure}
\affiliation{%
  \institution{Cornell University}
  \city{Ithaca}
  \state{NY}
  \postcode{14850}
  \country{USA}}
\email{hbc35@cornell.edu}



\renewcommand\shortauthors{Jung, M.F. et al}

\begin{abstract}
Research on human-robot collaboration or human-robot teaming, has focused predominantly on understanding and enabling collaboration between a single robot and a single human. Extending human-robot collaboration research beyond the dyad, raises novel questions about how a robot should distribute resources among group members and about what the social and task related consequences of the distribution are. Methodological advances are needed to allow researchers to collect data about human robot collaboration that involves multiple people. This paper presents Tower Construction, a novel resource distribution task that allows researchers to examine collaboration between a robot and groups of people. By focusing on the question of whether and how a robot's distribution of resources (wooden blocks required for a building task) affects collaboration dynamics and outcomes, we provide a case of how this task can be applied in a laboratory study with 124 participants to collect data about human robot collaboration that involves multiple humans. We highlight the kinds of insights the task can yield. In particular we find that the distribution of resources affects perceptions of performance, and interpersonal dynamics between human team-members. 
\end{abstract}

%
%

%
%


\maketitle

\section{Introduction}

\begin{figure}[h]
\centerline{\includegraphics[scale=0.215]{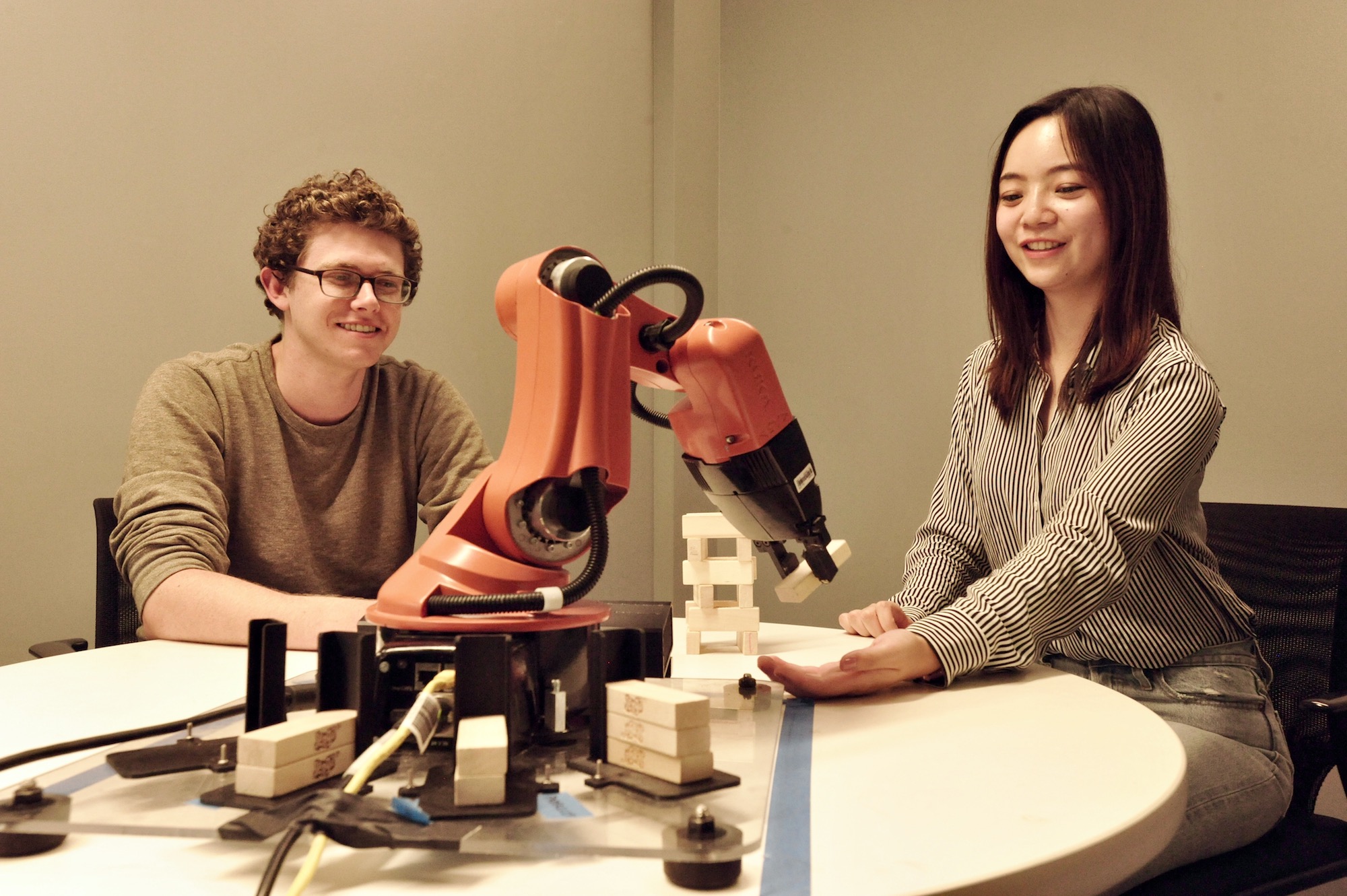}}
\caption{Two participants working as a team in completing a construction task in collaboration with a robot.}
\label{intropicture}
\end{figure}

\textit{A factory robot assists two workers by delivering parts needed for an engine assembly. One worker is experienced and fast, the other one is inexperienced and slow. The robot decides to assist the inexperienced worker.}

\vspace{3mm}

\noindent
\textit{An assistive robot in a stroke care facility is tasked to assist a group of patients in eating a meal. Two patients request the robot's assistance in picking up a piece of bread. The robot decides to help the older person.}

\vspace{3mm}

\noindent
\textit{An algorithm decides which employees in an office should receive a pay raise. It decides to give a pay raise to a recent hire.}

\vspace{3mm}

\noindent
\textit{Two customers, one foreign, enter a store. The robot greets the local customer first.}

\vspace{5mm}

\noindent
How would we want a robot to decide in each of these situations? What short and long-term consequences should we expect from a robot's behavior in these situations? Adding one or more humans to a human-robot dyad brings with it a dramatic change that current HRI theory does not capture, nor do current human-robot collaboration methods extend to such scenarios. Research on human-only groups has long recognized that adding just one more member to a dyad constitutes such fundamental change (e.g. \cite{simmel1964sociology} that researchers are still debating whether dyads should even be considered a group (e.g \cite{williams2010dyads, moreland2010dyads}). However, scenarios such as the ones above are becoming more common, as robots are increasingly deployed in complex social contexts that involve groups of people rather than individuals \cite{jung2018robots}. In order to build fundamental understanding and techniques that enable us to design robots that function well in group contexts, it is important to extend our understanding of human robot collaboration and robot assistance to group contexts. 

In the simplest of collaboration scenarios that involve multiple people, a robot has to make decisions about how to distribute resources (e.g. social attention, task support, or physical resources). A robot's distribution of resources likely affects the social dynamics within the group with potential effects on immediate and long term outcomes. For example, we know from prior work that people are uniquely attuned to inequalities and fairness in resource distribution  (e.g. \cite{brosnan2014evolution,lee2018understanding}). Simply distributing more of a robot's gaze towards one person in a group changes that person's role \cite{mutlu2009footing}. 

Extending the current human-robot collaboration paradigm toward multiple human collaborators raises a range of novel questions that cannot be answered by current human-robot collaboration approaches and theory. How should a robot distribute resources among multiple people? What impact does the manner of distribution have on people and their interpersonal dynamics? How does a robot's behavior affect how people interact with and relate to each other? How do people make sense of a robot's behavior in a group collaboration scenario? The answers to these questions can inform how we design algorithms that lead to improved task performance and favorable social dynamics.

To enable researchers to find answers to the types of questions listed above, we present a novel resource distribution task that allows researchers to collect data about human-robot collaboration that involves groups of people. Our task extends basic human-robot collaboration approaches (e.g. as in \cite{hayes2014online,hayes2015effective}) by adding one or more human collaborators. We provide a case of how this task can be applied in research by focusing on the question of how the unequal distribution of resources (wooden blocks required for a collaborative building task involving two humans) affects interpersonal collaboration dynamics and outcomes.

The paper makes three contributions to the HRI literature: First, we contribute a task that enables researchers to explore questions that arise when extending human-robot collaboration to groups of people. The task allows researchers to not only study how a robot's behavior impacts groups of people but it also offers a test-bed for the development and evaluation of novel algorithms that enable robots to interact and collaborate with multiple people. Our work constitutes a methodological contribution as it directly informs how research is carried out in HRI \cite{wobbrock2016research, wobbrock2012seven} by enabling researchers to collect data about human robot collaborations that involve more than one human. We demonstrate the utility of our task through a case study that explores the impact of unequal resource distribution on interpersonal dynamics and by highlighting a range of other possible use cases. Second, our task offers a context to collect data on social dynamics in human robot collaboration without requiring a robot that is designed with explicit social signaling capabilities (e.g. gaze, speech, or color displays).  While previous research has shown that people can make sense of any robot in social terms and thus recognize any machine as a social actor (e.g. \cite{reeves1996media, forlizzi2006service}), our task and case study extend this work by demonstrating that non-humanoid robots can also powerfully influence inter-human social dynamics. Third, our case study provides evidence that a robot distributing resources unequally between two human collaborators affects interpersonal dynamics in systematic ways. We specifically show that adding a second human to a human-robot collaboration scenario unavoidably requires the consideration of social dynamics between humans. In contrast to increasing work that investigates how robots can deliberately moderate the social dynamics of groups through the use of social robots, we show that not even the simplest of robot operations can be carried out without impacting the social configuration of the people involved.

\section{Studying Human-Robot Interaction in Groups and Teams}
To position our work within existing HRI research, we provide an overview of existing research on human-robot interaction in group and team settings. This review does not attempt to provide a comprehensive overview but rather aims at highlighting current trends and perspectives in HRI research on robots in group contexts.

\subsection{Human-Robot Collaboration and Teaming}
A growing amount of work on human-robot teaming has developed important insights toward understanding and enabling teamwork with embodied autonomous robots. Human-robot collaboration, or teaming, refers to collaborative partnerships between humans and robots in completing tasks and typically focuses on coordinating close, seamless joint activities between a single human and a single robot (for a recent review see \cite{ajoudani2018progress}). A central premise behind work to enable human-robot collaboration is that the combination of a robot's unique capabilities with those of humans opens possibilities that are out of reach for robots or humans each working in isolation (\cite{shah2017enhancing}). This involves a variety of behavioral and technical solutions for anticipating, planning, and executing partner actions (e.g. \cite{hayes2016autonomously, hoffman2007effects, shah2011improved}), generating interpretable behavior (e.g. \cite{dragan2013legibility, knepper2017implicit}, developing novel methods (e.g. crowdsourcing human behavior for model generation: \cite{breazeal2013crowdsourcing}), as well as studying effects of such collaborative partnerships on people (e.g. \cite{hinds2004whose}).  Several studies have also explored how human-robot teamwork can be improved through a robot's social behavior. For example, Scheutz, Schermerhorn, \& Kramer \cite{scheutz2006utility} found that team performance in a collaborative search task could be improved through emotional changes in the robot's voice tone, and several studies explored the important role of gaze in improving human-robot collaboration by studying handovers (e.g. \cite{admoni2014deliberate, moon2014meet}). Another study of a teamwork scenario in which one human participant interacted with two autonomous robots, found that robots can improve teamwork through subtle displays of positive engagement and interest alone \cite{jung2013engaging}.
While this work has been highly impactful in providing behavioral and computational insight into understanding and improving collaborative partnerships between a single human and a single robot, it does not, besides a few exceptions (e.g. \cite{strohkorb2018ripple, jung2015using, you2017emotional}), speak to contexts in which robots collaborate with more than a single person. 

\subsection{Robots in Work Contexts}

Studies of autonomous embodied robots and their effects on people in work contexts began as early as the 80s, soon after industrial robots saw increased use in manufacturing. For example, work by Linda Argote and colleagues (e.g. \cite{argote1985organizational, argote1983human}) demonstrated both advantages (reduced fatigue) and disadvantages (overabundance of human downtime) from the perspective of human workers. While these robots had to be employed in areas caged off from humans, recent advances in safe human-robot interaction have made it possible for humans to interact closely with robots without fear of, harm, injury, or death (see \cite{lasota2017survey} for a review) and robots now support work in close contact with humans in a wide range of areas. For example, a recent study by Sauppe and Mutlu \cite{sauppe2015social}  showed that a robot's treatment as a social entity extends to industrial settings in which robots closely work with humans on manufacturing tasks. Research on hospital delivery robots has demonstrated that how people respond to a robot depends on patient needs, work practices within a given unit \cite{mutlu2008robots}, and job roles \cite{siino2005robots}.
The introduction of robots can alter work practices and occupations. In a study of two pharmacy delivery robots in the UK, Barrett et al. \cite{barrett2012reconfiguring} found that pharmacy technicians expanded their role by tending to the robot whereas pharmacy assistants lost control over their work tasks, which were increasingly dictated by the demands of the robot and the technicians. Studies of science teams operating remote rovers also tell us that working with a remote autonomous robot can alter fundamental aspects of scientists' practices \cite{vertesi2015seeing} and that increased autonomy can lead to questions about how to interpret a robot's activities \cite{stubbs2007autonomy}.
In sum, studies of robots in work contexts have provided vital information about how robots change the way that people work, how people from different professions interact with robots, and how activities surrounding interactions with robots affect peoples' behavior and attitudes toward robots. However, we have little understanding about the mechanisms through which robots influence interpersonal dynamics in teams, despite decades of research on teams that establishes such dynamics as crucial for the short and long-term performance of teams.

\subsection{Robots in Group Settings}
A growing number of studies have demonstrated that embodied autonomous robots can affect their social environment beyond the persons interacting with them. Such influence can play out in multiple ways. We know that robots can take the role of a mediator and actively influence how two or more people relate to or interact with one another. Early studies of robots used in autism therapy, for example, show that robots can shape how children socially interact with others (for a review, see \cite{scassellati2012robots}). Further, Hoffman and colleagues presented a robot designed to diffuse tension in marital conflict through nonverbal fear expressions \cite{hoffman2015design} and a recent study by Shen, Slovak, \& Jung \cite{shen2018stop} demonstrated the effectiveness of a robot in helping young children resolve interpersonal resource conflicts. Expanding the focus from dyads to larger groups, Short and colleagues \cite{short2017understanding} showed that intergenerational interactions between older adults and their families could be shaped through a robot's behavior. Another study by Correia and colleagues \cite{correia2018group} showed that a group’s shared emotional state during a card playing game could be influenced through a robot's self or group oriented emotional statements. Several other studies have explored how the way a robot moves in a group affects coordination dynamics in groups engaged in group dance (e.g. \cite{iqbal2016movement, iqbal2017coordination}).
Studies have also begun to explore how groups of people are affected through subtle, often non-verbal behavior of a robot rather than through explicit moderation attempts. For example, Mutlu and colleagues \cite{mutlu2009footing} studied groups of museum visitors and demonstrated that even subtle, nonverbal gaze cues are sufficient to impact entire groups by shaping social roles. A study by Oliveira and colleagues \cite{oliveira2018friends} showed that the amount of pro-social behavior a group of people exhibited while playing a game was dependent on the robot's displayed goal orientation as friend or foe. Finally, a study by Vazquez and colleagues \cite{vazquez2017towards} showed that a robot's social orientation within a group of people affected social gaze patterns towards the robot.
Even robots that are not intentionally designed with a moderating role can shape interpersonal interactions in groups, for example, by shaping norms of social conduct through their behavior. Lee and colleagues documented "ripple effects" of a robot's behavior as it influenced how people interacted with each other and as it influenced the shared norms that people developed regarding appropriate ways of interacting with the robot \cite{lee2012ripple}. Longitudinal studies of the Robovie robot in shopping malls also show that people tend to approach a robot in groups and are more likely to act on the robot's advice about what stores to visit if they feel they have a relationship with it \cite{kanda2010communication}. There is also evidence that a robot's mere presence can affect how people interact with each other  \cite{dole2017dissertation, riether2012social}. Dole \cite{dole2017dissertation}, for example, showed that participants looked at a nutritionist more and perceived her as less warm when in the presence of a physically embodied social robot as compared to a tablet computer or a nonsocial robot.

These studies provide strong evidence that robots can influence people beyond the person interacting directly with the robot and thus shape how people behave and interact with others, the roles that people assume in groups, and even the norms that groups develop. However, despite this evidence for a robot's impact beyond the individual, most studies have been focusing on groups in family and play contexts (such as supporting family communication, or playing a game or dancing together) rather than on groups or teams in work or teamwork contexts. Additionally, all of the studies that demonstrated a group influence used robots with anthropomorphic features or robots that employed human behaviors such as speech.

\subsection{Summary}
Taken as a while, these studies highlight a growing interest in building understanding about a robot's influence on groups and about how robots can be deliberately designed with such group influence in mind. While many of these studies introduced novel collaborative tasks, these tasks are often (1) highly specific to the phenomenon under study (e.g. mistake admissions \cite{strohkorb2018ripple} or conflict reactions \cite{jung2015using}, (2) designed for anthropomorphic robots rather than standard robotic arms, and (3) have yet to explore whether and how non-anthropomorphic robots such as industrial robot arms can influence social dynamics in groups of people working together.
\section{Tower Construction -  A Resource Distribution Task}
To enable the extension of current human robot collaboration research towards groups of people collaborating with a robot, we developed a novel resource distribution task that extends typical human-robot collaboration scenarios beyond dyadic interactions. We designed our task to meet several criteria: First, the task should enable data-collection about human-robot interaction and  collaboration across a range of group sizes. That means, the method should be flexible and allow the same data collection approach when studying a robot interacting with two, three or more human collaborators. Second, the task should allow data collection of human-robot collaboration that employs a simple robot arm. Robot arms likely represent the most frequently deployed robots today, and robot arms are used across a broad range of human-robot collaboration contexts and studies (e.g. \cite{nikolaidis2013human,huang2015adaptive,wilcox2013optimization, dragan2013legibility}). Further, using a simple robotic arm capable of manipulating physical objects will allow us to draw insights that cannot be gained from virtual agents and physically embodied robots that lack manipulation capabilities (e.g. Nao robot, Jibo, or iCat). Third, the task should allow data collection about human robot collaboration across multiple levels of analysis (individual, inter-personal, and groups). The method should allow the collection of process and outcome data related to the task and social dynamics. Fourth, the task should be applicable to different forms of robot control. That means it should be applicable to robots that are Wizard of Oz controlled or autonomously controlled and thus allow data collection that 1) aids empirical studies aimed at understanding human robot collaboration, 2) aids the technical development and evaluation of novel algorithmic collaboration approaches, and 3) aids the generative design of novel interaction approaches. Fifth, the task should enable research that aims at developing novel understanding about a robot's impact on groups as well as research that aims at developing novel technical solutions (e.g. distribution algorithms) that enable robots to function in groups.

To meet these criteria we designed a task that places two or more human participants in a collaborative interaction with a robot. The robot decides how resources are distributed among participants. The task is specified through a task description, a physical setup, and a specification of a simple robot operation.

\subsection{Task Description}
The task requires a group of human participants to build a structure as tall as possible out of building blocks in collaboration with a robot. The robot's task is to deliver resources (building blocks) to the team by picking them up from a stack located behind the robot and placing them in the front, where participants are located. The task of the human participants is to place the blocks to build the desired structure. The robot decides how to distribute blocks. The task is completed once all blocks are placed.

This block building task models typical assembly or building tasks that are used in many other human-robot collaboration studies (e.g. \cite{khatib1999mobile,morioka2010new,hayes2014online,hayes2015effective}).

\subsection{Task Materials}
Task materials include a typical robot arm (e.g. Kuka YouBot) with a gripper as an end-effector, a set of building blocks, and a fixture to hold the blocks in place such that they can be reliably picked by the robot. The robot is placed on a table such that it can pick up building blocks from one side of the table and place them in front of the human collaborators on the other side of the table. Our example setup utilizes 20 blocks that are stored in four stacks of five blocks each.

The task utilizes a standard robot arm for three reasons. First, robot arms are possibly the most frequently employed type of robots due to their reliability, precision, and strength. Since its invention \cite{scheinman1969design} their use has been growing rapidly with an estimated 1.5$-$1.75 million industrial robots in operation in the US, and projections see a three-fold increase in use by 2025 \cite{acemoglu2017robots}. Second, advancements in safe human-robot interaction \cite{lasota2017survey} and human robot collaboration (e.g \cite{khatib1999mobile,morioka2010new,hayes2014online,hayes2015effective}) have increasingly placed robot arms into direct collaborative interaction with people further extending its use and making it increasingly important to understand how such robots affect their social environment. Third, focusing on the use of a robotic arm that places resources directly extends current research approaches to human-robot collaboration, or teaming, the majority of which employ similar robot arms (e.g. Khatib, 1999; Morioka \& Sakakibara, 2010; Hayes and Scassellati 2014b, 2015).

\begin{figure}[h!]
\centerline{\includegraphics[scale=0.13]{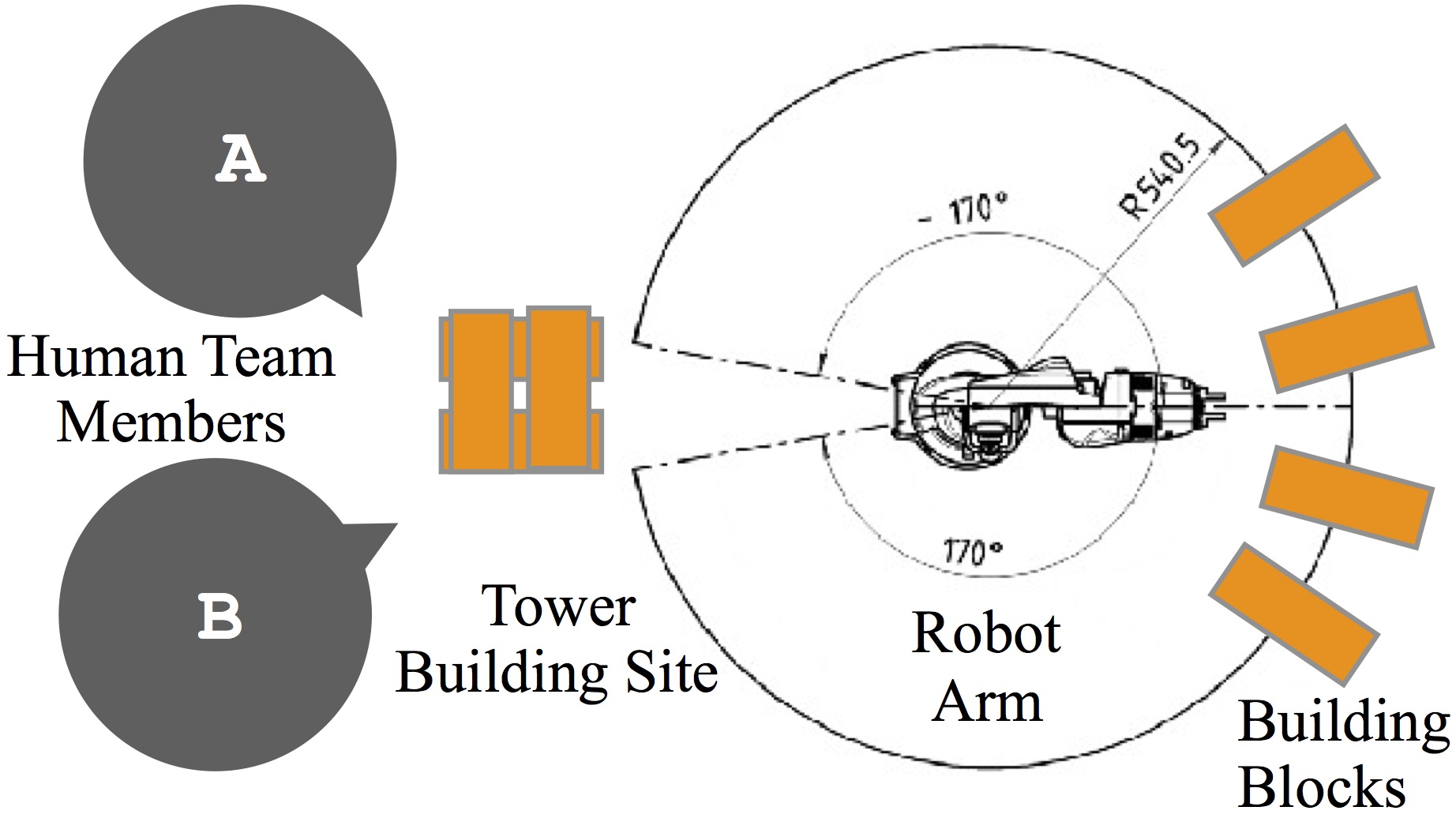}}
\caption{Schematic overview of the human-robot collaboration setup with two human collaborators as used in our case study. The orange squares represent building blocks that are picked up by the robot from the stacks on the right and then delivered to the participants who are using them to assemble a structure.}
\label{basic_setup}
\end{figure}

\subsection{Robot Operation}
In the simplest application of the task, the robot relies only on simple pick and place operations to collaborate on the task. Pick-and-place is an "elementary" robot operation that almost any robot arm can perform and that most robot arms do perform in one way or another across many current application contexts. By focusing on such elementary robot operations or behaviors, the method aims to aid in collecting data from which it is possible to draw more generalizable conclusions than is possible from data generated from more complex robot behaviors.
\section{Case Study: Applying the Task to a Study that Explores the Impact of Resource Distribution on Teamwork Outcomes}

Whenever a robot collaborates with more than one human it has to make decisions about the distribution of resources (e.g. social attention, task support, or physical resources). The robot has to decide to whom to attend, or whom to assist at any point in time. How such resources are distributed likely affects the social dynamics within the group with potential effects on immediate and long term team outcomes. For example, we know from prior work that people are uniquely attuned to inequalities and fairness in resource distribution (e.g. \cite{brosnan2014evolution,lee2018understanding}) and even a simple change in a robot’s distribution of gaze within a group of people can change participant roles \cite{mutlu2009footing}. Therefore, if we want to develop fundamental understanding about a robot’s impact on group behavior and outcomes it is important to understand whether and how resource distribution by a robot shapes group dynamics and outcomes.

To provide a case for the application of our task we used it to explore the question of how a robot's distribution of resources affects team social dynamics and outcomes in a collaborative construction task. Specifically we asked: Does the mere distribution of resources by a robot (equal vs unequal between two human team members) affect team performance and perceptions of relationship quality?

\subsection{Experimental Design}
To explore our research question, we designed an exploratory, two-condition (equal vs unequal distribution of resources), between participants study in which two human participants collaborate with a robot to complete a task - building a tower out of wooden blocks to be as high as possible. Our study directly extends current research on human-robot collaboration  (e.g. Khatib, 1999; Morioka \& Sakakibara, 2010; Hayes and Scassellati 2014b, 2015) by adding a second human team member. Equality in resource distribution (Independent Variable) was operationalized through the robot's placement of 20 wooden building blocks required to complete a tower building task. In the equal distribution condition, the robot gives 10 blocks to each participant, making a 50/50 distribution. In the unequal contribution condition, the robot gives the participant sitting in the right-hand side 13 blocks, and the participant on the left-hand 7 blocks, making a 65/35 distribution between the participants. Although we controlled the robot remotely using a Wizard of Oz approach, we introduced the robot as autonomous to participants and lead them to believe that the robot was actively watching and listening to their actions during the tower building task in order for it to make block distribution decisions.


\subsection{Participants}
We recruited 124 participants (N=62 teams) for this study using a combination of fliers and the university's on-line student recruitment system. Participants were provided compensation of either cash payments or student research credits. Of the 124 total participants recruited, 61.3 percent identified as female (n = 76), 37.9 percent were male (n = 47), and one participant identified as non-binary. A plurality of participants identified their ethnicity as Asian (42.7\%, n = 53), followed by White (33.9\%, n = 42), Hispanic (10.5\%, n = 13), Other (7.3\%, n = 9), African American (3.2\%, n = 4), and Pacific Islander (0.8\%, n = 1), with two participants choosing not to identify their ethnicity. Ages ranged considerably from 18 - 66 years, but the median age of participants was 21 years. Given this study occurred on a university campus and relied largely on an undergraduate and graduate participant pool, most participants were highly educated with over 75 percent (n = 94) reporting having at least some college education. 

\subsection{Procedure}
The study procedures involved four steps totaling between 30 and 45 minutes.

\textit{Step 1 - Pre Survey and Task Introduction:} Participants first provided informed consent and filled out a pre-study survey (data not included in this case study) that included demographic questions, questions about their relationship status with their study partner, and their experience with AI and robots in general. Next, an experimenter led participants into a small laboratory where the robot was set up as described in the previous section. The experimenter asked participants to sit on one of the randomly assigned chairs.

Once seated, the experimenter told participants that the study involved testing an autonomous robot designed to help a two-person team build the tallest tower possible from a set of 20 wooden blocks. The experiment would consist of 20 rounds (one for each wooden block). At the start of each round, the robotic arm would give one of the participants a wooden block. That participant must use this block and place it to construct a tower in front of the robot. Once the block had been placed, the robotic arm would start the next round by picking up a new block and deciding which team member to give it to. This would continue for twenty rounds until all the blocks had been used. Participants were told that their goal was to build a tower as tall as they could. They were free to talk and plan with one another, but they could only place a block on the tower if it was given to them by the robot (participant were explicitly told they could not share blocks with each other).

\begin{figure}[h!]
\centerline{\includegraphics[scale=0.382]{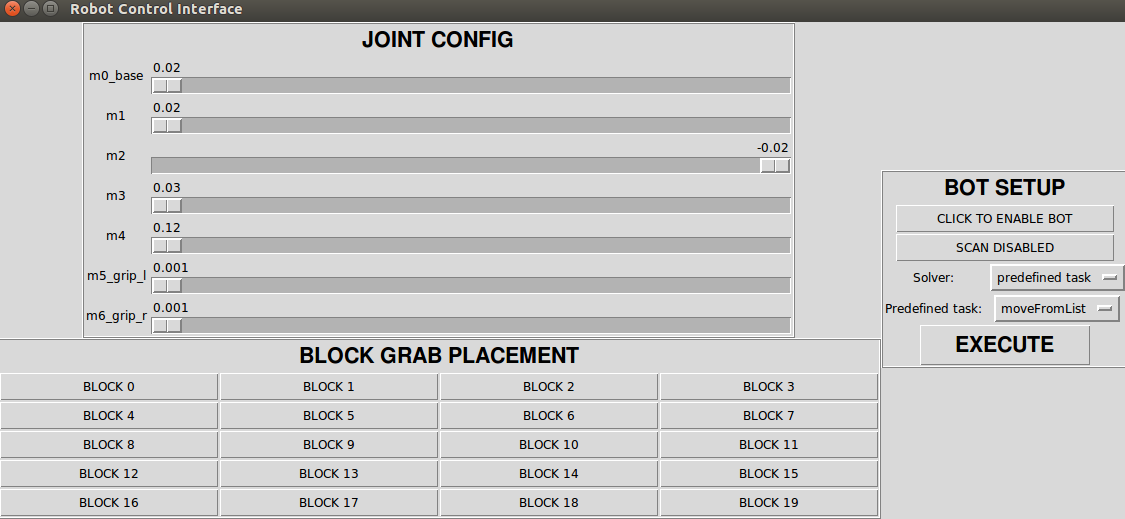}}
\caption{Remote operator interface for Wizard of Oz control of the robot}
\label{remoteinterface}
\end{figure}

To highlight the interdependent nature of the task, we informed participants that each team could earn an extra team bonus of up to \$10 dependent on the height of the tower measured at the end of the building task. However, the team bonus would be paid out according to individual task contribution and split up according to the distribution of block placements. If a team earned a total bonus of \$8 and each participant placed 10 blocks on the tower, that would be a 50/50 split, and each participant would receive \$4. Thus, across conditions participants were told that the bonus compensation would depend on the team's success and that the bonus would be distributed according to the distribution of block placements. If the participants' tower fell down at anytime during the study, the table would be cleared, and the participants would continue building the tower with whatever blocks remained. They were reminded that this would significantly lower their bonus payment, as their tower would have fewer blocks. We used this reward design to model teamwork contexts in which both team success and individual contributions contribute to rewards and to highlight the interdependent nature of the collaboration task.

\textit{Step 2 - Calibration Routine:} The second step of the procedure was designed to create a believable experience of working with an autonomous robot. While the robot was operated by the experimenter through a remote operation interface (see figure \ref{remoteinterface}), it was introduced to participants as autonomous. The experimenter told participants that a machine learning approach had been used to train the robot on a dataset of hundreds of two person teams building similar wooden block towers. Based on this training, the robot would give a block to the person it thought at the moment could best help the team succeed towards building the highest tower.

To further emphasize the seemingly autonomous nature of the robot, we asked participants to complete a short calibration and training routine with the robot before starting the actual task. Participants were told that the robot has two main sensors, a Microsoft Kinect camera placed in front of the robot that is able to map and model the participants movements in 3D space, and a microphone. Both visual and audio information would be used by the robot to determine to whom it should give the next block. In actuality, none of the systems explained here were functional but were merely placed on the table. The first part of the calibration routine required each participant to pick up a wooden block three times with each hand in front of the robot. Participants were told that this data was used to build a 3D model of their movements. The second part of the calibration routine involved each participant uttering "Mary had a little lamb who's fleece was white as snow" 3 times in their normal speaking voice. Participants were told this was used to build a vocal model for the robot to understand their speech. In both cases, the experimenter looked at a laptop seemingly entering parameters and monitoring the progress of the calibration routine until it was completed in a satisfactory fashion.



\begin{table}[]
\centering
\caption{Order robot hands out blocks in each round for each condition.}
\label{block_table}
\resizebox{\columnwidth}{!}{%
\setlength{\tabcolsep}{0.5em} 
{\renewcommand{\arraystretch}{1.5}
\begin{tabular}{|l|l|l|l|l|l|l|l|l|l|l|l|l|l|l|l|l|l|l|l|l|}
\hline
Condition & 1 & 2 & 3 & 4 & 5 & 6 & 7 & 8 & 9 & 10 & 11 & 12 & 13 & 14 & 15 & 16 & 17 & 18 & 19 & 20 \\ \hline
65/35 & A & B & A & B & A & A & A & B & A & B & A & A & A & B & A & B & A & A & A & B \\ \hline
50/50 & A & B & A & B & A & A & B & B & A & B & A & A & B & B & A & B & A & A & B & B \\ \hline
\end{tabular}
}}
\end{table}

\textit{Step 3 - Building Task:} After the completion of the calibration routine, the experimenter left the room telling participants that he/she would wait outside the room during the tower building activity. Instead, the experiment monitored a live video feed of the task in the nearby observation lab.

Participants were asked to start with the building task once the robot handed out the first block. Once a participant placed a block on the tower, the robot continued by passing out the next block. This continued until all 20 blocks were passed out to the participants. Based on the study condition, blocks were distributed in the sequences outlined in Table \ref{block_table} (A,B = participants A and B as in figure \ref{basic_setup}).

At the completion of the task, the experimenter entered the room, measured the height of the tower with a ruler. The experimenter told all participants, irrespective of their actual performance, that their tower was in the 89th percentile of all towers measured so far and would thus receive 8/10 points, equaling a team bonus of \$8. Based on the study condition and the distribution of blocks, participants were then ether paid both \$4 in the equal distribution (50/50) condition, or \$6 and \$2 respectively in the unequal distribution (65/35) condition.

\textit{Step 4 - Post Survey and Debriefing} After receiving their reward payments, participants completed a series of survey measures including the subjective value inventory and open ended questions asking them to reflect on the tower building activity and the robot's decision-making mechanism. Participants were then debriefed and told the true purpose of the study, that the robot was not actually watching or listen to them, and that the distribution they experienced was pre-programmed before they started the study.

\subsection{Measures}
To evaluate the impact of the robot's behavior on the team we measured team performance and team members' satisfaction with the quality of the relationship with their team partner.

\subsubsection{Team Performance}
Our team performance measure was operationalized by collecting a measure of each team's tower height at the conclusion of the building task. An objective measure of overall team performance was analyzed both to establish whether distribution conditions significantly impacted task performance and, if so, to statistically control for the effects of these differences on how team members subjectively experience the task.

\subsubsection{Relationship Satisfaction} 
We adapted items from a subscale (Feelings About Counterpart Relations) of Subjective Value Inventory (SVI) from Curhan et al \cite{curhan2006people} to understand participants satisfaction with their relationship with their team partner. All items were measured on 7-point Likert scales (1 = strongly disagree to 7 = strongly agree) and had high reliability ($\alpha$=.82). The items measured the participants' general impression of their partner and the effect of the activity on their sense of trust and satisfaction (e.g., "Did the activity make you trust your counterpart?").  

\subsection{Results}
\begin{figure}[h!]
\centerline{\includegraphics[scale=0.30]{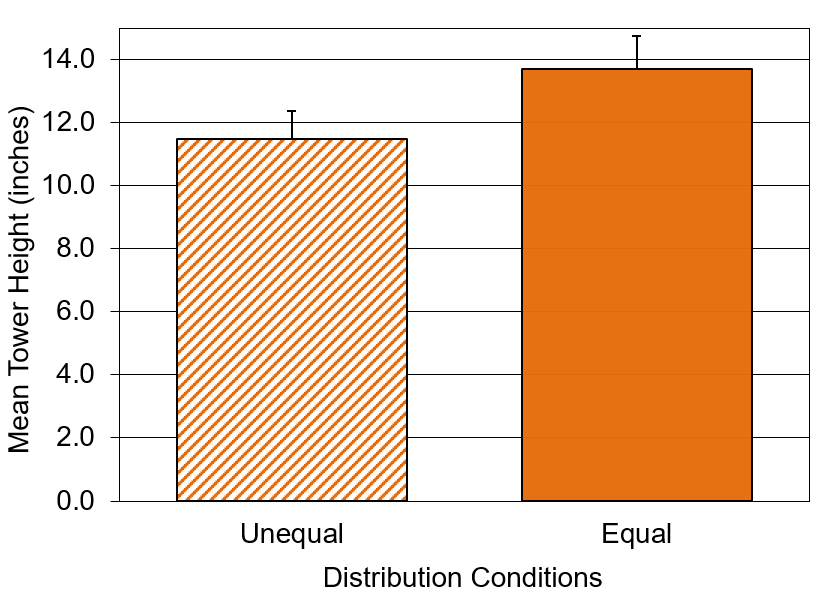}\includegraphics[scale=0.30]{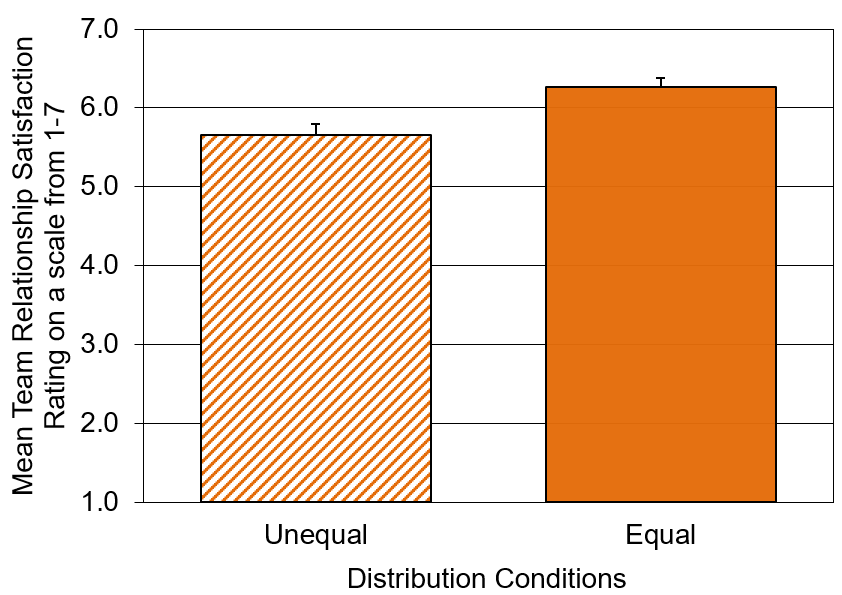}}
\caption{Left: Mean tower height between distribution conditions. Not significant at the $p< .05$ level. Right: Mean ratings of team relationship satisfaction between distribution conditions. Significant at the $p< .01$ level. Error bars represent the standard error.}
\label{towerheightgraph}
\end{figure}

In order to test whether the KUKA Youbot, through mere differences in block distribution (equal vs. unequal distribution), was capable of influencing team member ratings of relationship satisfaction, a group-level relationship satisfaction score was calculated by averaging individual team member ratings within each team. We conducted a single independent samples t-test using the team relationship satisfaction score as the dependent variable. Levene's test of equality of variance was non-significant indicating that equality of variance could be assumed, F(1,60)=1.78, p $=$ .19. Results of the Student's t-test demonstrate a significant difference between distribution conditions, t(1,60)=-2.86, p $=$ .006, d $=$ .73, such that teams in the unequal distribution condition reported less team relationship satisfaction than those in the equal distribution condition (see table \ref{dvdescriptives}). Tower height, as an objective team performance metric, was not controlled in the statistical analysis of team relationship satisfaction, because it did not differ significantly between conditions t(1,60) = 1.63, p $=$ .11 (see figure \ref{towerheightgraph}).


\begin{table}[h]
	\centering
	\caption{Descriptive statistics of team relationship satisfaction for each condition}
	\label{dvdescriptives}
	\begin{tabular}{lrrrrr}
		\hline
		 & Block Distribution & N & Mean & SD & SE  \\
		\hline
		Team Relationship Satisfaction & Unequal & 30 & 5.646 & 0.757 & 0.138  \\
		  & Equal & 32 & 6.160 & 0.659 & 0.117  \\
		\hline
	\end{tabular} 
\end{table}

\begin{table}[h]
	\centering
	\caption{Independent Samples T-Test of effects of robot block distribution on team relationship satisfaction}
	\begin{tabular}{lrrrrrr}
		\hline
		\multicolumn{1}{c}{} & \multicolumn{1}{c}{} & \multicolumn{1}{c}{} & \multicolumn{1}{c}{} & \multicolumn{1}{c}{} & \multicolumn{2}{c}{95\% CI for Cohen's d} \\
		\cline{6-7}
		 & t & df & p & Cohen's d & Lower & Upper  \\
		\hline
		Team Relationship Satisfaction & -2.857 & 60.00 & 0.006 & -0.726 & -1.238 & -0.209  \\
		\hline
	\end{tabular} 
\end{table}

\subsection{Discussion}
Results from this case study demonstrate the impact that a non-anthropomorphic robot can have on team outcomes simply by the way it decides to distribute resources among team members. Team members in the unequal distribution condition reported a significantly more negative perception of their relationship than those in the equal distribution condition. 

Our finding that the distribution of resources by a robot affects the team's overall relationship quality has implications for both design practice and theory. First, this study informs how we might design algorithms to assist groups of people. Our study demonstrates that the development of algorithms for robots that support team in achieving optimal performance might face a difficult dilemma. On the one hand a robot's assistance should help the team optimize overall task performance by distributing resources to the team member most capable of helping the team succeed. On the other hand, our study showed that distributing resources unequally can come at a social cost which might hurt team performance in the long term as studies of teams in organizations have highlighted the importance of relationship quality and cohesion for performance (e.g. \cite{cohen1997makes}). Thus when designing algorithms for robots to assist teams, researchers face the challenge to find ways that optimize immediate task performance while also maintaining good social dynamics between members of the team.

Our finding that unequal resource distribution leads to an impoverished interpersonal relationship perception  has implications for our understanding of algorithmic resource distribution and fairness. Recent research on algorithmic resource distribution between multiple people has predominantly focused on understanding individuals' perceptions of the algorithm and its decisions (e.g. \cite{lee2017algorithmic, lee2018understanding}). Less is known about how resource distribution and perception of those as fair or unfair affects peoples' interpersonal relationships with each other. Our study sheds light onto the possible negative interpersonal consequences that may arise from letting machines decide how to distribute resources. 

Our also provided initial evidence that sheds light onto the complex social role an industrial robot can take on when engaged in a group interaction. Table \ref{handover_table} shows a 19 second sequence from one of the teams participating in the study. One of the participants (left participant)) places her hand open on the table in expectation of receiving the next block. Then, seemingly unexpected to her, the robot places the block in front of the other participant (on the right). This creates a vulnerable moment for the participant on the left, who makes a tense facial expression and retracts her arm. Upon receiving the block, the participant on the right displays a nervous frown, seemingly embarrassed about the position the robot put him in. The sequence ends with with an exchange of gaze and affiliative smiles between participants as part of what seems an attempt to repair the awkward social situation the robot has placed them in. This exchange is interesting for several reasons: It demonstrates not how a robot's simple pick and place behavior can become offensive as it places one participant in a vulnerable position through a seeming rejection. Further, the example demonstrates how a robot's simple behavior can impact interpersonal dynamics between multiple people. The awkward situation created by the robot makes a repair relevant. Finally, this example shows how even an industrial robot without any anthropomorphic physical features and without making a gesture that is specifically designed for social legibility (e.g. as in \cite{takayama2011expressing, dragan2013legibility}) can elicit complex social dynamics within a group through one of the most basic robot operations an industrial robot can perform: picking up and placing an object. As such this work builds on early findings on peoples' social relations to non-anthropomorphic robots (e.g. \cite{forlizzi2006service}, by showing that non-anthropomorphic robots can also shape peoples' social relations with each other in complex ways.

Finally our case study provided initial evidence about people's reasoning about a robot's decision making in groups. We asked people about how they thought the robot makes resource distribution decisions and we received a wide range of answers. For example, some participants thought resource distribution decisions depended on the robot's assessment of individual building skills: "I believe the robots actions were determined by looking at the way we placed the blocks in terms of location but as well as how we placed the blocks like gentle, firm, or rigid. Because of my gentleness I believe the robot chose me to place many of the blocks in order to prevent falls and build a higher tower." Other participants believed the robot's decisions were driven by participants' verbal utterances: "The robot seemed to judge it mainly at first based on what we said, since I stated [sic] not having done this before. But later on it started taking into account our ideas such as different way of stacking the blocks. I think it did a good job being responsive to our ideas." Thus, this task can help in collecting data about how people reason about a robot's decision making in group situations and contribute to ongoing debates on the lay theories people form when interacting with algorithmic systems (e.g. \cite{devito2018algorithm}). 

\par
\noindent
\begin{table}[]
\centering
\caption{Short interaction sequence evolving around a robot's block placement.}
\label{handover_table}
\begin{tabu} to \textwidth {@{} X[l]  X[l] }
 \includegraphics[width= 0.48\textwidth]{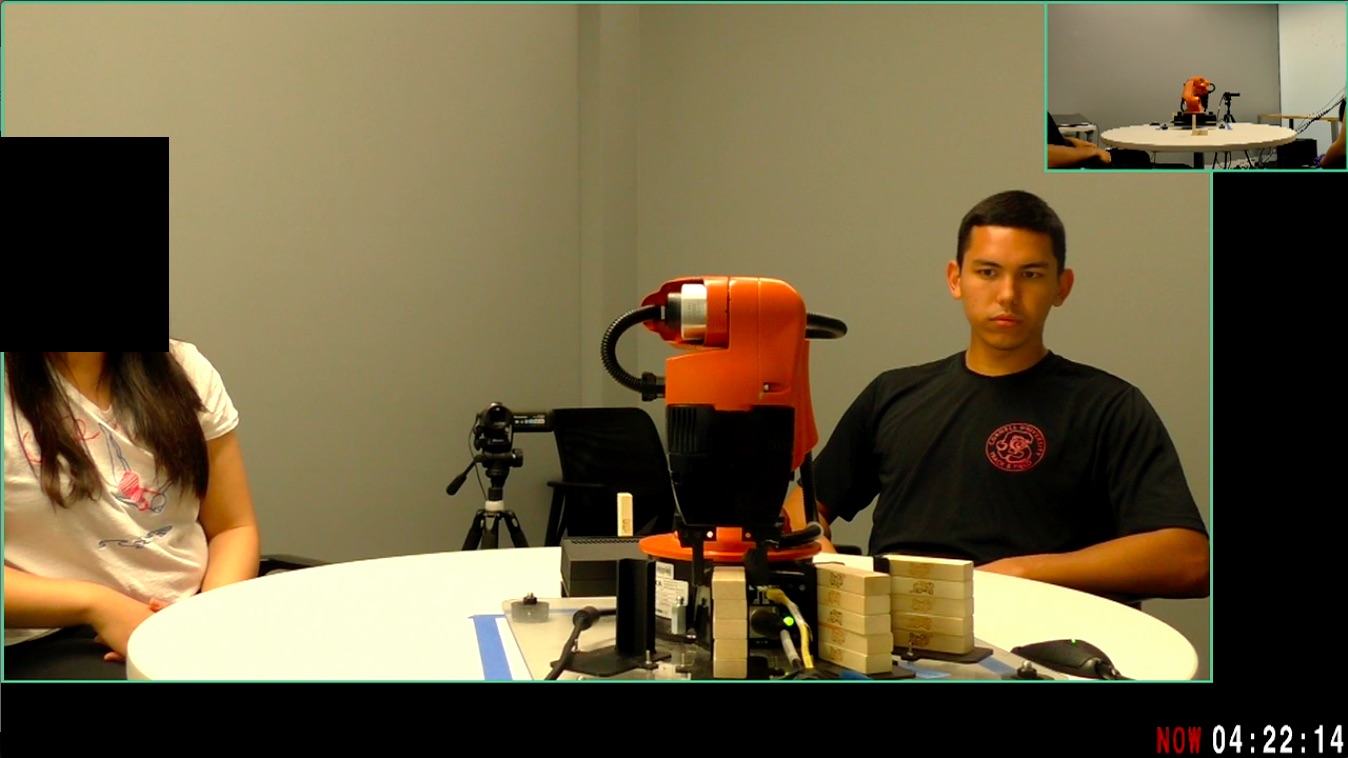} & \includegraphics[width= 0.48\textwidth]{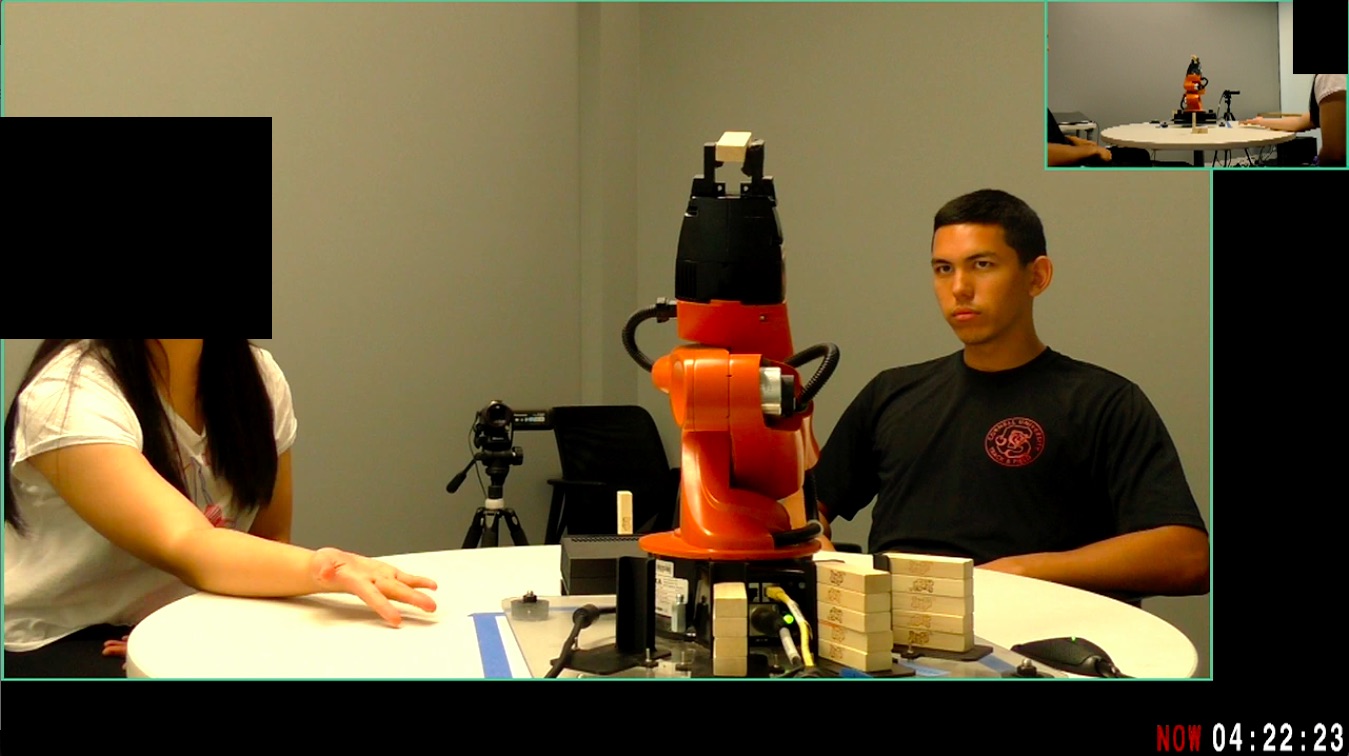} \\
 \textit{1) 00 seconds}: Participants are waiting for the robot to pick up the next block.  &
 \textit{2) 09 seconds}: The participant on the left extends her arm in expectation of receiving the next block \\
 \includegraphics[width= 0.48\textwidth]{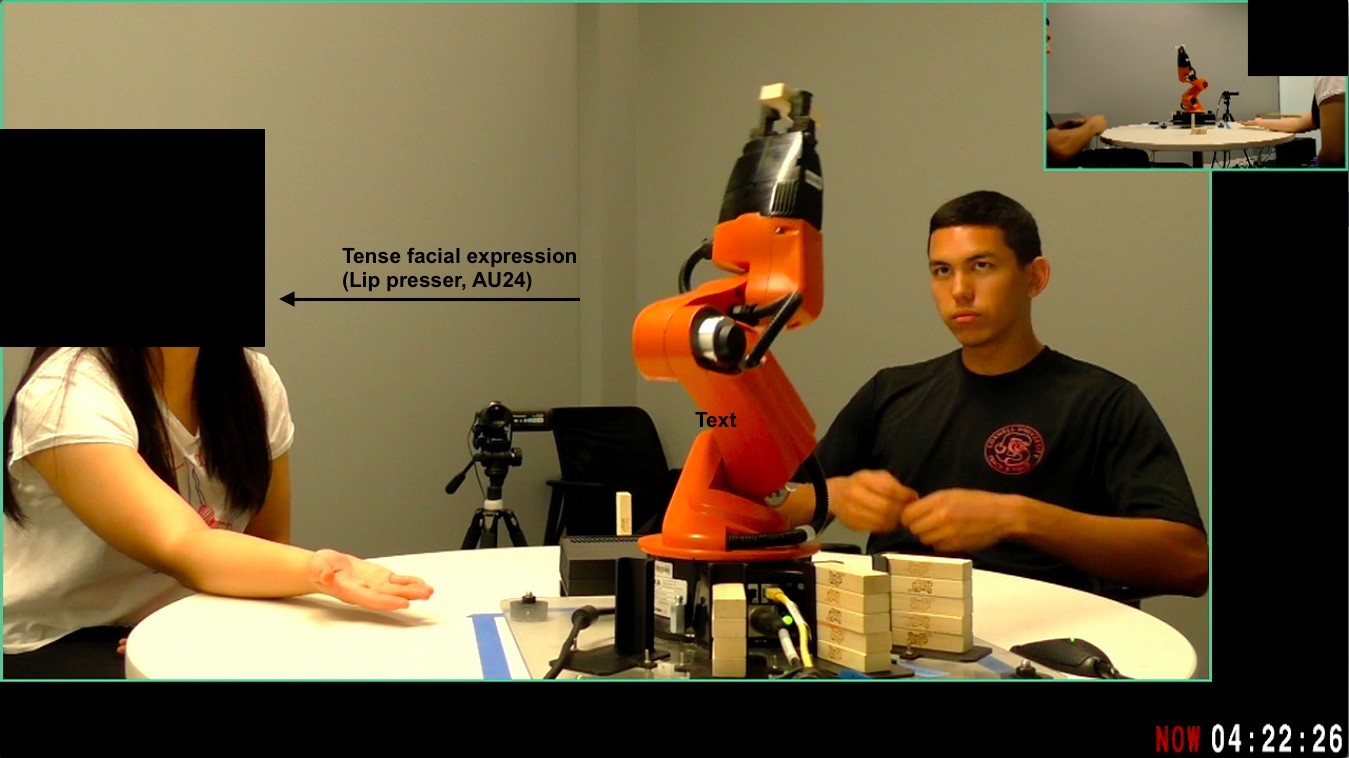} & \includegraphics[width= 0.48\textwidth]{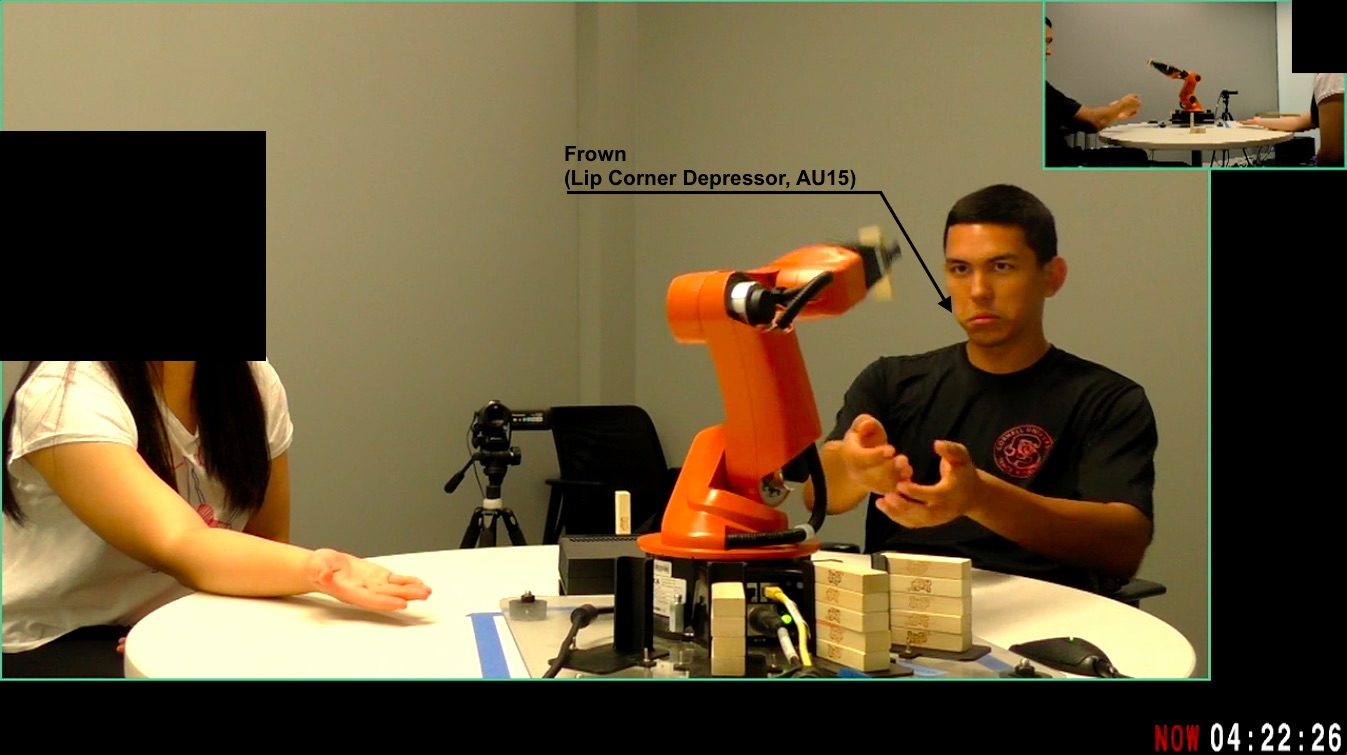} \\
 \textit{3) 12 seconds}: Upon realizing that the robot decided to give the next block to the participant on the right, the participant on the left makes a tense facial expression (lip presser, AU24) reflecting the vulnerable position the robot left her in &
 \textit{4) 12 seconds}: Upon receiving the next block, the participant on the right makes a tense expression of discomfort (lip corner depressor, AU15) which reflects the awkwardness of the situation the participants were placed in by the robot \\
 \includegraphics[width= 0.48\textwidth]{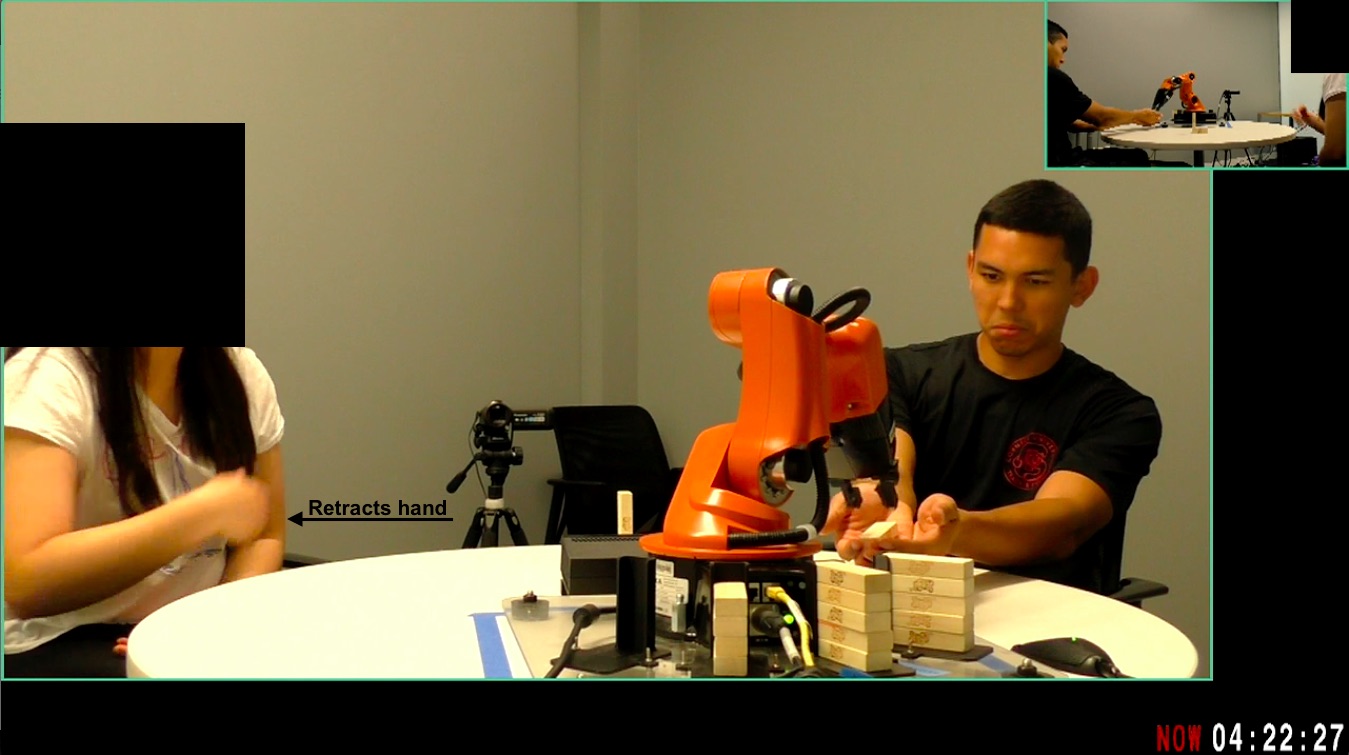} & \includegraphics[width= 0.48\textwidth]{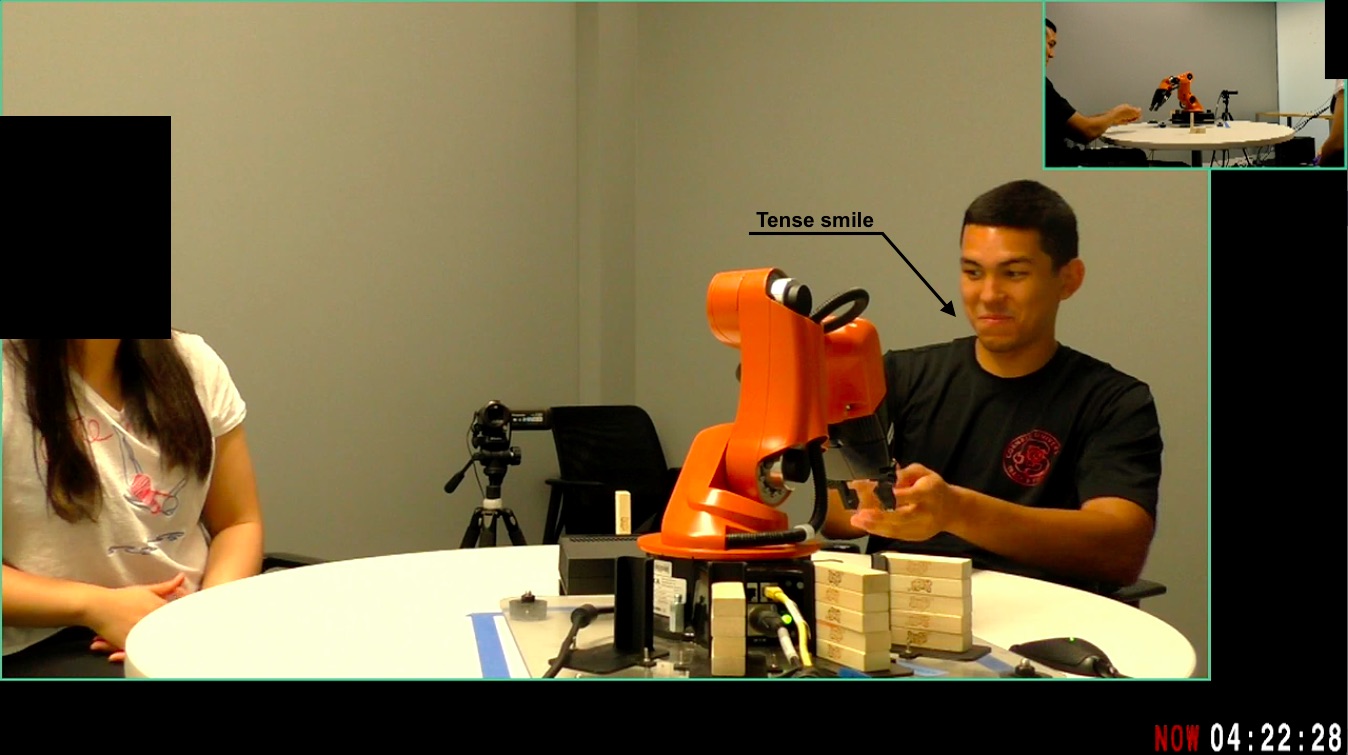} \\
 \textit{5) 13 seconds}: The participant on the left awkwardly retracts her hand as the other participant receives the block &
 \textit{6) 14 seconds}: The participant on the right expresses an awkward smile towards the participant on the left in an attempt to repair the situation's awkwardness  \\
 \includegraphics[width= 0.48\textwidth]{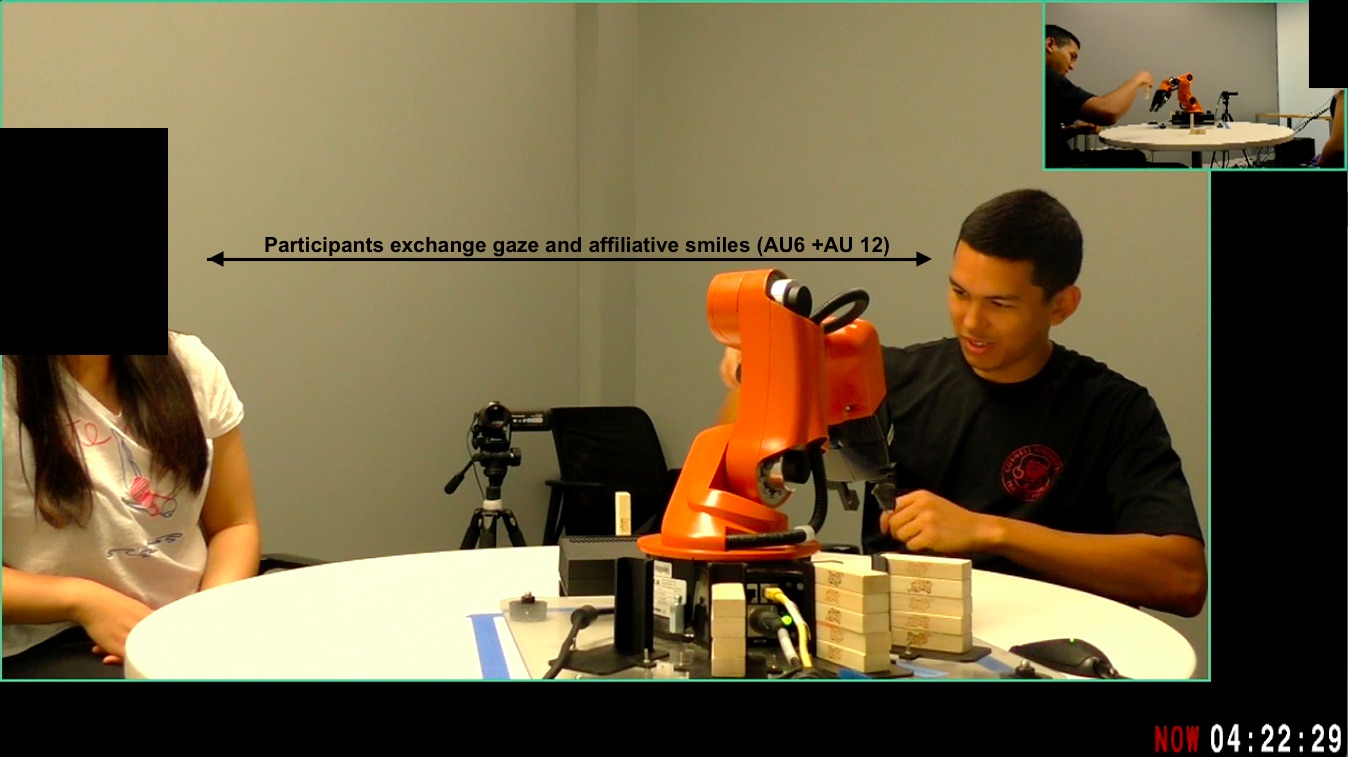} & \includegraphics[width= 0.48\textwidth]{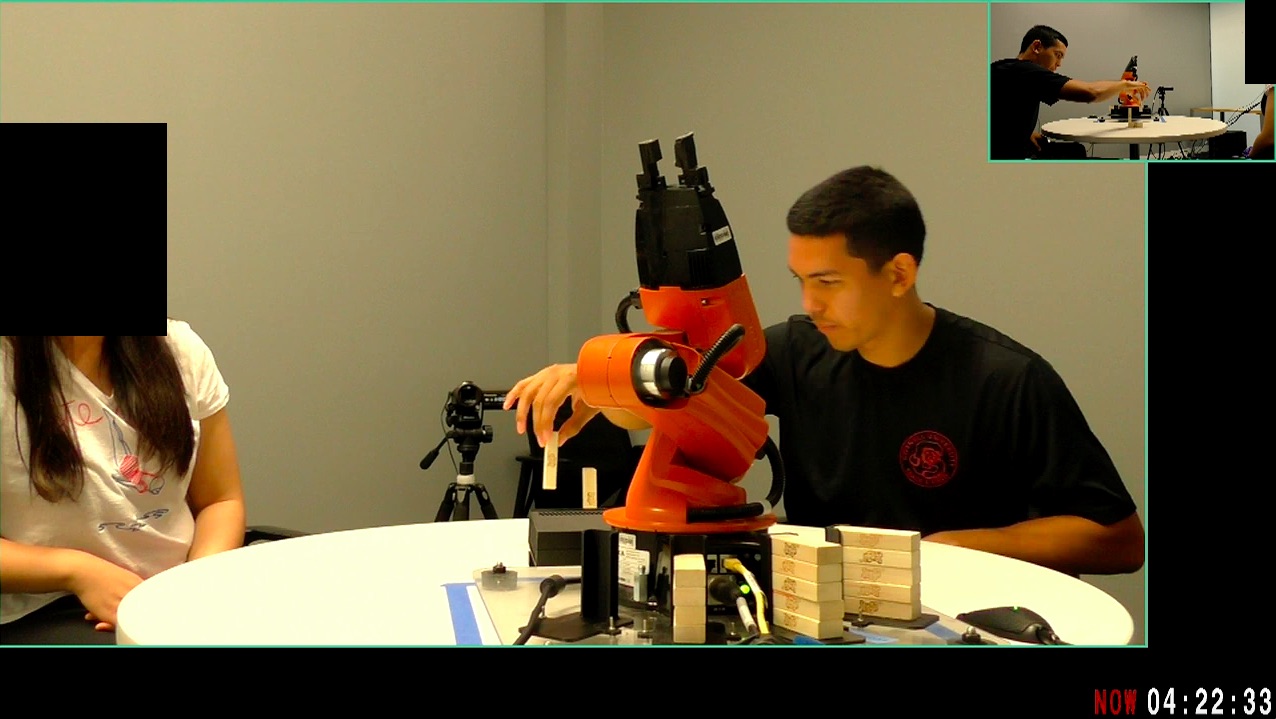}\\
 \textit{7) 15 seconds}: Both participants express affiliative smiles (lip corner puller, AU12 and cheek raiser AU6) &
 \textit{8) 19 seconds}: The participant on the right places the received as the other participant is watching\\
\end{tabu}
\end{table}
\section{Overall Discussion and Conclusion}

We introduced a Resource Distribution Task that enables HRI researchers to collect data about human-robot collaboration that involves multiple human participants. The task engages a group of people in a collaborative interaction with a robot that distributes resources among group members.

We demonstrated the utility of our approach through a case study that shows how this task enabled us to collect data on the impact of a robot's resource distribution through elementary pick and place behaviors on interpersonal dynamics and outcomes. Our work directly informs how research in HRI is carried out \cite{wobbrock2016research, wobbrock2012seven} as it enables researchers to collect data about human robot collaboration that involves more than one human. More broadly, our paper contributes to a growing effort of building HRI specific research methodology (e.g.\cite{dautenhahn2007methodology, bethel2010review, woods2006methodological,riek2012wizard,steinfeld2006common, hoffman2014designing}).

\subsection{Advantages over existing tasks}
The resource distribution task is, of course, not the first task that generally engages a single robot in an interaction with multiple people. Other tasks exist, they are are often difficult to replicate as they rely on highly complex study protocols that involve a highly trained confederate (e.g. \cite{jung2015using}, a complex game designed to explore one specific research question (e.g. \cite{strohkorb2018ripple, oliveira2018friends, short2017understanding}), a specific robot design (e.g. \cite{hoffman2014designing}), or they focus on remote controlled rather than autonomous robots (e.g. \cite{you2017emotional}). To our understanding, the resource distribution task is the first task that allows the examination of human-robot collaboration and teaming situations that involve multiple people. While many human-robot teaming and collaboration tasks exist they all involve single robot in a collaborative interaction with a single human (e.g. \cite{khatib1999mobile,morioka2010new,hayes2014online,hayes2015effective}). 
 
\begin{figure*}[h!t]
\centering
\includegraphics[width=\textwidth]{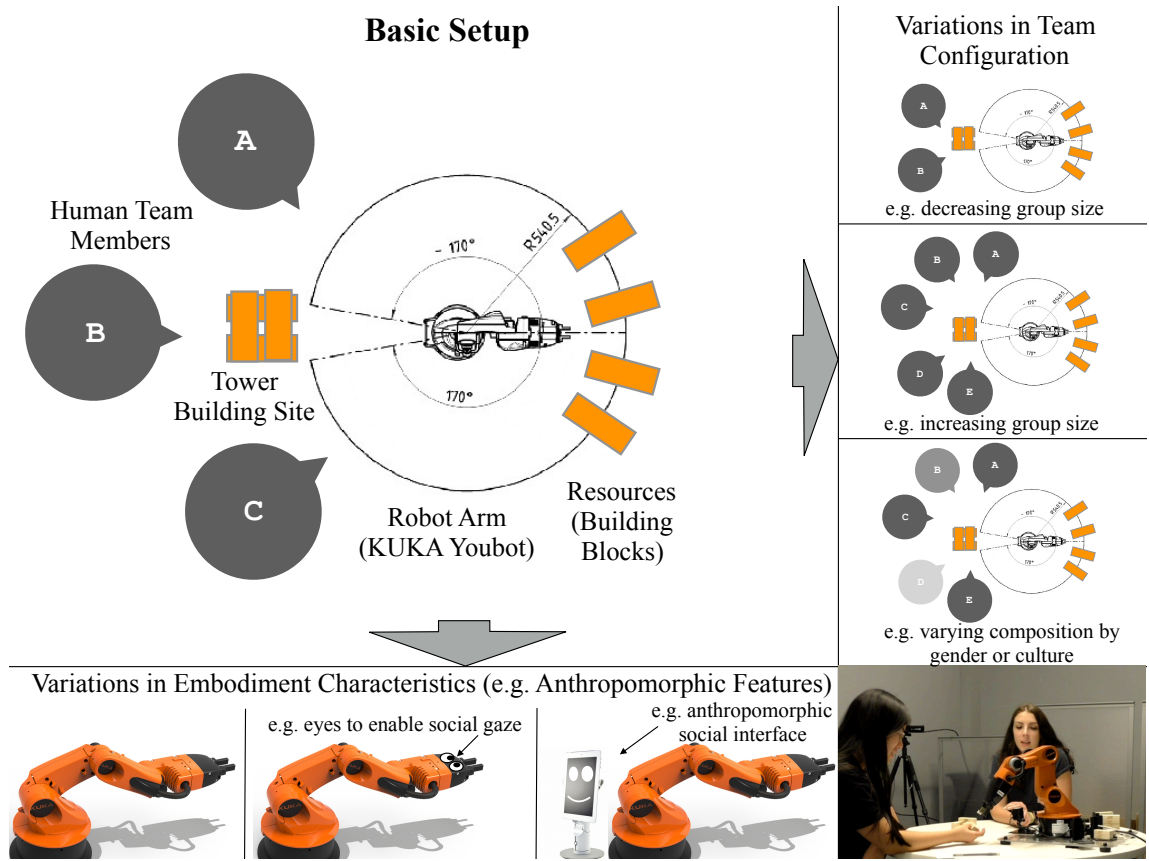}
\caption{Basic setup and possible alternate uses by modifying the ask along dimensions of group size and composition as well as anthropomorphic design features of the robot}
\label{extendedsetup}
\end{figure*}
 
A particular strength of this task is that it can be easily modified. For example, team size or team composition (e.g. by gender or culture) can be easily modified to study effects of a robot's collaborative behavior on group size and composition(see figure \ref{extendedsetup}. Further, it is possible to study varying degrees of anthropomorphic embodiment, for example, by adding eyes to study gaze effects (e.g. as in \cite{admoni2014deliberate, moon2014meet}), or to study different motion trajectories when handing over blocks (e.g. \cite{dragan2013legibility}), or by adding more complex social interfaces through a screen as is common for many current human safe robot arms (e.g. iRobot's Baxter and Sawyer, ABB's Yumi, or KUKA's LBR iiwa).

\subsection{Generalizability}
While the task to collaborate with a robot arm to construct a tower out of wooden blocks seems very specific, we believe this task can speak to a broad range of human-robot interaction scenarios. As we laid out in the introduction, questions regarding resource distribution emerge whenever robots are deployed in group settings. Our task provides a tool to examine questions surrounding a robot's resource distribution in group settings. Just as the Desert Survival Task \cite{lafferty1974desert} does not only speak to communication in a desert survival situation, and just as the famous ostracism task \cite{williams1997social} does not only speak to exclusion effects during a ball game, our task is intended to speak to any situation in which a robot distributes a resource among members of a group.  

Beyond studying how unequal resource distribution affects interpersonal dynamics and outcomes in groups, we believe that this task lends itself to a broad range of research questions in the area of human-robot collaboration. As one application for the task, we envision it to be used in the design and evaluation of novel algorithms for human robot collaboration in groups. For example, it could be used to evaluate resource distribution algorithms and test their impact on team performance and interpersonal dynamics. A second application context for this task could be as a scenario for designing novel interaction strategies for robots in group settings, using a generative wizard of oz paradigm (e.g. \cite{hoffman2014designing}). A third application  also help in developing understanding how people make sense of robots and how people reason about a robot's decision making. As a final application contest, the task might also be used to explore the use a robot's resource distribution as a deliberate intervention into a group's social dynamics. For example, the robot could provide more resources to the person that is not actively contributing to the task or less resources to the person who dominates the floor in order to create more equal group participation.

\subsection{Conclusion}
More than a century ago, Simmel reflected that adding a third member to a dyad brings with it a dramatic qualitative change as "the triad exhibits in its simplest form the sociological drama that informs all social life: the dialectic of freedom and constraint, of autonomy and heteronomy" \cite{coser1971masters}:187. Adding just one more person to dyadic human-robot collaboration brings with it complexities that are yet to be fully understood. We hope this task enables others to expand our understanding of human-robot collaboration research beyond the dyad.

\bibliographystyle{ACM-Reference-Format}
\bibliography{bibliography}


\begin{thebibliography}{72}


\ifx \showCODEN    \undefined \def \showCODEN     #1{\unskip}     \fi
\ifx \showDOI      \undefined \def \showDOI       #1{#1}\fi
\ifx \showISBNx    \undefined \def \showISBNx     #1{\unskip}     \fi
\ifx \showISBNxiii \undefined \def \showISBNxiii  #1{\unskip}     \fi
\ifx \showISSN     \undefined \def \showISSN      #1{\unskip}     \fi
\ifx \showLCCN     \undefined \def \showLCCN      #1{\unskip}     \fi
\ifx \shownote     \undefined \def \shownote      #1{#1}          \fi
\ifx \showarticletitle \undefined \def \showarticletitle #1{#1}   \fi
\ifx \showURL      \undefined \def \showURL       {\relax}        \fi
\providecommand\bibfield[2]{#2}
\providecommand\bibinfo[2]{#2}
\providecommand\natexlab[1]{#1}
\providecommand\showeprint[2][]{arXiv:#2}

\bibitem[\protect\citeauthoryear{Acemoglu and Restrepo}{Acemoglu and
  Restrepo}{2017}]%
        {acemoglu2017robots}
\bibfield{author}{\bibinfo{person}{Daron Acemoglu} {and}
  \bibinfo{person}{Pascual Restrepo}.} \bibinfo{year}{2017}\natexlab{}.
\newblock \showarticletitle{Robots and jobs: Evidence from US labor markets}.
\newblock  (\bibinfo{year}{2017}).
\newblock


\bibitem[\protect\citeauthoryear{Admoni, Dragan, Srinivasa, and
  Scassellati}{Admoni et~al\mbox{.}}{2014}]%
        {admoni2014deliberate}
\bibfield{author}{\bibinfo{person}{Henny Admoni}, \bibinfo{person}{Anca
  Dragan}, \bibinfo{person}{Siddhartha~S Srinivasa}, {and}
  \bibinfo{person}{Brian Scassellati}.} \bibinfo{year}{2014}\natexlab{}.
\newblock \showarticletitle{Deliberate delays during robot-to-human handovers
  improve compliance with gaze communication}. In
  \bibinfo{booktitle}{\emph{Proceedings of the 2014 ACM/IEEE international
  conference on Human-robot interaction}}. ACM, \bibinfo{pages}{49--56}.
\newblock


\bibitem[\protect\citeauthoryear{Ajoudani, Zanchettin, Ivaldi,
  Albu-Sch{\"a}ffer, Kosuge, and Khatib}{Ajoudani et~al\mbox{.}}{2018}]%
        {ajoudani2018progress}
\bibfield{author}{\bibinfo{person}{Arash Ajoudani},
  \bibinfo{person}{Andrea~Maria Zanchettin}, \bibinfo{person}{Serena Ivaldi},
  \bibinfo{person}{Alin Albu-Sch{\"a}ffer}, \bibinfo{person}{Kazuhiro Kosuge},
  {and} \bibinfo{person}{Oussama Khatib}.} \bibinfo{year}{2018}\natexlab{}.
\newblock \showarticletitle{Progress and prospects of the human--robot
  collaboration}.
\newblock \bibinfo{journal}{\emph{Autonomous Robots}} (\bibinfo{year}{2018}),
  \bibinfo{pages}{1--19}.
\newblock


\bibitem[\protect\citeauthoryear{Argote and Goodman}{Argote and
  Goodman}{1985}]%
        {argote1985organizational}
\bibfield{author}{\bibinfo{person}{Linda Argote} {and} \bibinfo{person}{Paul~S
  Goodman}.} \bibinfo{year}{1985}\natexlab{}.
\newblock \showarticletitle{The organizational implications of robotics}.
\newblock  (\bibinfo{year}{1985}).
\newblock


\bibitem[\protect\citeauthoryear{Argote, Goodman, and Schkade}{Argote
  et~al\mbox{.}}{1983}]%
        {argote1983human}
\bibfield{author}{\bibinfo{person}{Linda Argote}, \bibinfo{person}{Paul~S
  Goodman}, {and} \bibinfo{person}{David Schkade}.}
  \bibinfo{year}{1983}\natexlab{}.
\newblock \showarticletitle{The Human Side of Robotics: How Worker's React to a
  Robot}.
\newblock \bibinfo{journal}{\emph{Sloan Management Review}}
  (\bibinfo{year}{1983}).
\newblock


\bibitem[\protect\citeauthoryear{Barrett, Oborn, Orlikowski, and Yates}{Barrett
  et~al\mbox{.}}{2012}]%
        {barrett2012reconfiguring}
\bibfield{author}{\bibinfo{person}{Michael Barrett}, \bibinfo{person}{Eivor
  Oborn}, \bibinfo{person}{Wanda~J Orlikowski}, {and} \bibinfo{person}{JoAnne
  Yates}.} \bibinfo{year}{2012}\natexlab{}.
\newblock \showarticletitle{Reconfiguring boundary relations: Robotic
  innovations in pharmacy work}.
\newblock \bibinfo{journal}{\emph{Organization Science}} \bibinfo{volume}{23},
  \bibinfo{number}{5} (\bibinfo{year}{2012}), \bibinfo{pages}{1448--1466}.
\newblock


\bibitem[\protect\citeauthoryear{Bethel and Murphy}{Bethel and Murphy}{2010}]%
        {bethel2010review}
\bibfield{author}{\bibinfo{person}{Cindy~L Bethel} {and}
  \bibinfo{person}{Robin~R Murphy}.} \bibinfo{year}{2010}\natexlab{}.
\newblock \showarticletitle{Review of human studies methods in HRI and
  recommendations}.
\newblock \bibinfo{journal}{\emph{International Journal of Social Robotics}}
  \bibinfo{volume}{2}, \bibinfo{number}{4} (\bibinfo{year}{2010}),
  \bibinfo{pages}{347--359}.
\newblock


\bibitem[\protect\citeauthoryear{Breazeal, DePalma, Orkin, Chernova, and
  Jung}{Breazeal et~al\mbox{.}}{2013}]%
        {breazeal2013crowdsourcing}
\bibfield{author}{\bibinfo{person}{Cynthia Breazeal}, \bibinfo{person}{Nick
  DePalma}, \bibinfo{person}{Jeff Orkin}, \bibinfo{person}{Sonia Chernova},
  {and} \bibinfo{person}{Malte Jung}.} \bibinfo{year}{2013}\natexlab{}.
\newblock \showarticletitle{Crowdsourcing human-robot interaction: New methods
  and system evaluation in a public environment}.
\newblock \bibinfo{journal}{\emph{Journal of Human-Robot Interaction}}
  \bibinfo{volume}{2}, \bibinfo{number}{1} (\bibinfo{year}{2013}),
  \bibinfo{pages}{82--111}.
\newblock


\bibitem[\protect\citeauthoryear{Brosnan and de~Waal}{Brosnan and
  de~Waal}{2014}]%
        {brosnan2014evolution}
\bibfield{author}{\bibinfo{person}{Sarah~F Brosnan} {and}
  \bibinfo{person}{Frans~BM de Waal}.} \bibinfo{year}{2014}\natexlab{}.
\newblock \showarticletitle{Evolution of responses to (un) fairness}.
\newblock \bibinfo{journal}{\emph{Science}} \bibinfo{volume}{346},
  \bibinfo{number}{6207} (\bibinfo{year}{2014}), \bibinfo{pages}{1251776}.
\newblock


\bibitem[\protect\citeauthoryear{Cohen and Bailey}{Cohen and Bailey}{1997}]%
        {cohen1997makes}
\bibfield{author}{\bibinfo{person}{Susan~G Cohen} {and}
  \bibinfo{person}{Diane~E Bailey}.} \bibinfo{year}{1997}\natexlab{}.
\newblock \showarticletitle{What makes teams work: Group effectiveness research
  from the shop floor to the executive suite}.
\newblock \bibinfo{journal}{\emph{Journal of management}} \bibinfo{volume}{23},
  \bibinfo{number}{3} (\bibinfo{year}{1997}), \bibinfo{pages}{239--290}.
\newblock


\bibitem[\protect\citeauthoryear{Correia, Mascarenhas, Prada, Melo, and
  Paiva}{Correia et~al\mbox{.}}{2018}]%
        {correia2018group}
\bibfield{author}{\bibinfo{person}{Filipa Correia}, \bibinfo{person}{Samuel
  Mascarenhas}, \bibinfo{person}{Rui Prada}, \bibinfo{person}{Francisco~S
  Melo}, {and} \bibinfo{person}{Ana Paiva}.} \bibinfo{year}{2018}\natexlab{}.
\newblock \showarticletitle{Group-based emotions in teams of humans and
  robots}. In \bibinfo{booktitle}{\emph{Proceedings of the 2018 ACM/IEEE
  International Conference on Human-Robot Interaction}}. ACM,
  \bibinfo{pages}{261--269}.
\newblock


\bibitem[\protect\citeauthoryear{Coser}{Coser}{1971}]%
        {coser1971masters}
\bibfield{author}{\bibinfo{person}{Lewis~A Coser}.}
  \bibinfo{year}{1971}\natexlab{}.
\newblock \bibinfo{booktitle}{\emph{Masters of sociological thought: Ideas in
  historical and social context}}.
\newblock \bibinfo{publisher}{Houghton Mifflin Harcourt P}.
\newblock


\bibitem[\protect\citeauthoryear{Curhan, Elfenbein, and Xu}{Curhan
  et~al\mbox{.}}{2006}]%
        {curhan2006people}
\bibfield{author}{\bibinfo{person}{Jared~R Curhan},
  \bibinfo{person}{Hillary~Anger Elfenbein}, {and} \bibinfo{person}{Heng Xu}.}
  \bibinfo{year}{2006}\natexlab{}.
\newblock \showarticletitle{What do people value when they negotiate? Mapping
  the domain of subjective value in negotiation.}
\newblock \bibinfo{journal}{\emph{Journal of personality and social
  psychology}} \bibinfo{volume}{91}, \bibinfo{number}{3}
  (\bibinfo{year}{2006}), \bibinfo{pages}{493}.
\newblock


\bibitem[\protect\citeauthoryear{Dautenhahn}{Dautenhahn}{2007}]%
        {dautenhahn2007methodology}
\bibfield{author}{\bibinfo{person}{Kerstin Dautenhahn}.}
  \bibinfo{year}{2007}\natexlab{}.
\newblock \showarticletitle{Methodology \& themes of human-robot interaction: A
  growing research field}.
\newblock \bibinfo{journal}{\emph{International Journal of Advanced Robotic
  Systems}} \bibinfo{volume}{4}, \bibinfo{number}{1} (\bibinfo{year}{2007}),
  \bibinfo{pages}{15}.
\newblock


\bibitem[\protect\citeauthoryear{DeVito, Hancock, French, Birnholtz, Antin,
  Karahalios, Tong, and Shklovski}{DeVito et~al\mbox{.}}{2018}]%
        {devito2018algorithm}
\bibfield{author}{\bibinfo{person}{Michael~A DeVito},
  \bibinfo{person}{Jeffrey~T Hancock}, \bibinfo{person}{Megan French},
  \bibinfo{person}{Jeremy Birnholtz}, \bibinfo{person}{Judd Antin},
  \bibinfo{person}{Karrie Karahalios}, \bibinfo{person}{Stephanie Tong}, {and}
  \bibinfo{person}{Irina Shklovski}.} \bibinfo{year}{2018}\natexlab{}.
\newblock \showarticletitle{The Algorithm and the User: How Can HCI Use Lay
  Understandings of Algorithmic Systems?}. In
  \bibinfo{booktitle}{\emph{Extended Abstracts of the 2018 CHI Conference on
  Human Factors in Computing Systems}}. ACM, \bibinfo{pages}{panel04}.
\newblock


\bibitem[\protect\citeauthoryear{Dole}{Dole}{2017}]%
        {dole2017dissertation}
\bibfield{author}{\bibinfo{person}{Lorin Dole}.}
  \bibinfo{year}{2017}\natexlab{}.
\newblock \showarticletitle{The influence of a robot’s mere presence on human
  communication}.
\newblock \bibinfo{journal}{\emph{PhD Dissertation}} (\bibinfo{year}{2017}).
\newblock


\bibitem[\protect\citeauthoryear{Dragan, Lee, and Srinivasa}{Dragan
  et~al\mbox{.}}{2013}]%
        {dragan2013legibility}
\bibfield{author}{\bibinfo{person}{Anca~D Dragan}, \bibinfo{person}{Kenton~CT
  Lee}, {and} \bibinfo{person}{Siddhartha~S Srinivasa}.}
  \bibinfo{year}{2013}\natexlab{}.
\newblock \showarticletitle{Legibility and predictability of robot motion}. In
  \bibinfo{booktitle}{\emph{Proceedings of the 8th ACM/IEEE international
  conference on Human-robot interaction}}. IEEE Press,
  \bibinfo{pages}{301--308}.
\newblock


\bibitem[\protect\citeauthoryear{Forlizzi and DiSalvo}{Forlizzi and
  DiSalvo}{2006}]%
        {forlizzi2006service}
\bibfield{author}{\bibinfo{person}{Jodi Forlizzi} {and} \bibinfo{person}{Carl
  DiSalvo}.} \bibinfo{year}{2006}\natexlab{}.
\newblock \showarticletitle{Service robots in the domestic environment: a study
  of the roomba vacuum in the home}. In \bibinfo{booktitle}{\emph{Proceedings
  of the 1st ACM SIGCHI/SIGART conference on Human-robot interaction}}. ACM,
  \bibinfo{pages}{258--265}.
\newblock


\bibitem[\protect\citeauthoryear{Hayes and Scassellati}{Hayes and
  Scassellati}{2014}]%
        {hayes2014online}
\bibfield{author}{\bibinfo{person}{Bradley Hayes} {and} \bibinfo{person}{Brian
  Scassellati}.} \bibinfo{year}{2014}\natexlab{}.
\newblock \showarticletitle{Online development of assistive robot behaviors for
  collaborative manipulation and human-robot teamwork}. In
  \bibinfo{booktitle}{\emph{Proceedings of the” Machine Learning for
  Interactive Systems”(MLIS) Workshop at AAAI}}.
\newblock


\bibitem[\protect\citeauthoryear{Hayes and Scassellati}{Hayes and
  Scassellati}{2015}]%
        {hayes2015effective}
\bibfield{author}{\bibinfo{person}{Bradley Hayes} {and} \bibinfo{person}{Brian
  Scassellati}.} \bibinfo{year}{2015}\natexlab{}.
\newblock \showarticletitle{Effective robot teammate behaviors for supporting
  sequential manipulation tasks}. In \bibinfo{booktitle}{\emph{Intelligent
  Robots and Systems (IROS), 2015 IEEE/RSJ International Conference on}}. IEEE,
  \bibinfo{pages}{6374--6380}.
\newblock


\bibitem[\protect\citeauthoryear{Hayes and Scassellati}{Hayes and
  Scassellati}{2016}]%
        {hayes2016autonomously}
\bibfield{author}{\bibinfo{person}{Bradley Hayes} {and} \bibinfo{person}{Brian
  Scassellati}.} \bibinfo{year}{2016}\natexlab{}.
\newblock \showarticletitle{Autonomously constructing hierarchical task
  networks for planning and human-robot collaboration}. In
  \bibinfo{booktitle}{\emph{Robotics and Automation (ICRA), 2016 IEEE
  International Conference on}}. IEEE, \bibinfo{pages}{5469--5476}.
\newblock


\bibitem[\protect\citeauthoryear{Hinds, Roberts, and Jones}{Hinds
  et~al\mbox{.}}{2004}]%
        {hinds2004whose}
\bibfield{author}{\bibinfo{person}{Pamela~J Hinds}, \bibinfo{person}{Teresa~L
  Roberts}, {and} \bibinfo{person}{Hank Jones}.}
  \bibinfo{year}{2004}\natexlab{}.
\newblock \showarticletitle{Whose job is it anyway? A study of human-robot
  interaction in a collaborative task}.
\newblock \bibinfo{journal}{\emph{Human-Computer Interaction}}
  \bibinfo{volume}{19}, \bibinfo{number}{1} (\bibinfo{year}{2004}),
  \bibinfo{pages}{151--181}.
\newblock


\bibitem[\protect\citeauthoryear{Hoffman and Breazeal}{Hoffman and
  Breazeal}{2007}]%
        {hoffman2007effects}
\bibfield{author}{\bibinfo{person}{Guy Hoffman} {and} \bibinfo{person}{Cynthia
  Breazeal}.} \bibinfo{year}{2007}\natexlab{}.
\newblock \showarticletitle{Effects of anticipatory action on human-robot
  teamwork efficiency, fluency, and perception of team}. In
  \bibinfo{booktitle}{\emph{Proceedings of the ACM/IEEE international
  conference on Human-robot interaction}}. ACM, \bibinfo{pages}{1--8}.
\newblock


\bibitem[\protect\citeauthoryear{Hoffman and Ju}{Hoffman and Ju}{2014}]%
        {hoffman2014designing}
\bibfield{author}{\bibinfo{person}{Guy Hoffman} {and} \bibinfo{person}{Wendy
  Ju}.} \bibinfo{year}{2014}\natexlab{}.
\newblock \showarticletitle{Designing robots with movement in mind}.
\newblock \bibinfo{journal}{\emph{Journal of Human-Robot Interaction}}
  \bibinfo{volume}{3}, \bibinfo{number}{1} (\bibinfo{year}{2014}),
  \bibinfo{pages}{91--122}.
\newblock


\bibitem[\protect\citeauthoryear{Hoffman, Zuckerman, Hirschberger, Luria, and
  Shani~Sherman}{Hoffman et~al\mbox{.}}{2015}]%
        {hoffman2015design}
\bibfield{author}{\bibinfo{person}{Guy Hoffman}, \bibinfo{person}{Oren
  Zuckerman}, \bibinfo{person}{Gilad Hirschberger}, \bibinfo{person}{Michal
  Luria}, {and} \bibinfo{person}{Tal Shani~Sherman}.}
  \bibinfo{year}{2015}\natexlab{}.
\newblock \showarticletitle{Design and evaluation of a peripheral robotic
  conversation companion}. In \bibinfo{booktitle}{\emph{Proceedings of the
  Tenth Annual ACM/IEEE International Conference on Human-Robot Interaction}}.
  ACM, \bibinfo{pages}{3--10}.
\newblock


\bibitem[\protect\citeauthoryear{Huang, Cakmak, and Mutlu}{Huang
  et~al\mbox{.}}{2015}]%
        {huang2015adaptive}
\bibfield{author}{\bibinfo{person}{Chien-Ming Huang}, \bibinfo{person}{Maya
  Cakmak}, {and} \bibinfo{person}{Bilge Mutlu}.}
  \bibinfo{year}{2015}\natexlab{}.
\newblock \showarticletitle{Adaptive Coordination Strategies for Human-Robot
  Handovers.}. In \bibinfo{booktitle}{\emph{Robotics: Science and Systems}}.
\newblock


\bibitem[\protect\citeauthoryear{Iqbal, Rack, and Riek}{Iqbal
  et~al\mbox{.}}{2016}]%
        {iqbal2016movement}
\bibfield{author}{\bibinfo{person}{Tariq Iqbal}, \bibinfo{person}{Samantha
  Rack}, {and} \bibinfo{person}{Laurel~D Riek}.}
  \bibinfo{year}{2016}\natexlab{}.
\newblock \showarticletitle{Movement coordination in human--robot teams: a
  dynamical systems approach}.
\newblock \bibinfo{journal}{\emph{IEEE Transactions on Robotics}}
  \bibinfo{volume}{32}, \bibinfo{number}{4} (\bibinfo{year}{2016}),
  \bibinfo{pages}{909--919}.
\newblock


\bibitem[\protect\citeauthoryear{Iqbal and Riek}{Iqbal and Riek}{2017}]%
        {iqbal2017coordination}
\bibfield{author}{\bibinfo{person}{Tariq Iqbal} {and} \bibinfo{person}{Laurel~D
  Riek}.} \bibinfo{year}{2017}\natexlab{}.
\newblock \showarticletitle{Coordination dynamics in multihuman multirobot
  teams}.
\newblock \bibinfo{journal}{\emph{IEEE Robotics and Automation Letters}}
  \bibinfo{volume}{2}, \bibinfo{number}{3} (\bibinfo{year}{2017}),
  \bibinfo{pages}{1712--1717}.
\newblock


\bibitem[\protect\citeauthoryear{Jung and Hinds}{Jung and Hinds}{2018}]%
        {jung2018robots}
\bibfield{author}{\bibinfo{person}{Malte Jung} {and} \bibinfo{person}{Pamela
  Hinds}.} \bibinfo{year}{2018}\natexlab{}.
\newblock \showarticletitle{Robots in the Wild: A Time for More Robust Theories
  of Human-Robot Interaction}.
\newblock \bibinfo{journal}{\emph{ACM Transactions on Human-Robot Interaction
  (THRI)}} \bibinfo{volume}{7}, \bibinfo{number}{1} (\bibinfo{year}{2018}),
  \bibinfo{pages}{2}.
\newblock


\bibitem[\protect\citeauthoryear{Jung, Lee, DePalma, Adalgeirsson, Hinds, and
  Breazeal}{Jung et~al\mbox{.}}{2013}]%
        {jung2013engaging}
\bibfield{author}{\bibinfo{person}{Malte~F Jung}, \bibinfo{person}{Jin~Joo
  Lee}, \bibinfo{person}{Nick DePalma}, \bibinfo{person}{Sigurdur~O
  Adalgeirsson}, \bibinfo{person}{Pamela~J Hinds}, {and}
  \bibinfo{person}{Cynthia Breazeal}.} \bibinfo{year}{2013}\natexlab{}.
\newblock \showarticletitle{Engaging robots: easing complex human-robot
  teamwork using backchanneling}. In \bibinfo{booktitle}{\emph{Proceedings of
  the 2013 conference on Computer supported cooperative work}}. ACM,
  \bibinfo{pages}{1555--1566}.
\newblock


\bibitem[\protect\citeauthoryear{Jung, Martelaro, and Hinds}{Jung
  et~al\mbox{.}}{2015}]%
        {jung2015using}
\bibfield{author}{\bibinfo{person}{Malte~F Jung}, \bibinfo{person}{Nikolas
  Martelaro}, {and} \bibinfo{person}{Pamela~J Hinds}.}
  \bibinfo{year}{2015}\natexlab{}.
\newblock \showarticletitle{Using robots to moderate team conflict: the case of
  repairing violations}. In \bibinfo{booktitle}{\emph{Proceedings of the Tenth
  Annual ACM/IEEE International Conference on Human-Robot Interaction}}. ACM,
  \bibinfo{pages}{229--236}.
\newblock


\bibitem[\protect\citeauthoryear{Kanda, Shiomi, Miyashita, Ishiguro, and
  Hagita}{Kanda et~al\mbox{.}}{2010}]%
        {kanda2010communication}
\bibfield{author}{\bibinfo{person}{Takayuki Kanda}, \bibinfo{person}{Masahiro
  Shiomi}, \bibinfo{person}{Zenta Miyashita}, \bibinfo{person}{Hiroshi
  Ishiguro}, {and} \bibinfo{person}{Norihiro Hagita}.}
  \bibinfo{year}{2010}\natexlab{}.
\newblock \showarticletitle{A communication robot in a shopping mall}.
\newblock \bibinfo{journal}{\emph{IEEE Transactions on Robotics}}
  \bibinfo{volume}{26}, \bibinfo{number}{5} (\bibinfo{year}{2010}),
  \bibinfo{pages}{897--913}.
\newblock


\bibitem[\protect\citeauthoryear{Khatib}{Khatib}{1999}]%
        {khatib1999mobile}
\bibfield{author}{\bibinfo{person}{Oussama Khatib}.}
  \bibinfo{year}{1999}\natexlab{}.
\newblock \showarticletitle{Mobile manipulation: The robotic assistant}.
\newblock \bibinfo{journal}{\emph{Robotics and Autonomous Systems}}
  \bibinfo{volume}{26}, \bibinfo{number}{2-3} (\bibinfo{year}{1999}),
  \bibinfo{pages}{175--183}.
\newblock


\bibitem[\protect\citeauthoryear{Knepper, Mavrogiannis, Proft, and
  Liang}{Knepper et~al\mbox{.}}{2017}]%
        {knepper2017implicit}
\bibfield{author}{\bibinfo{person}{Ross~A Knepper},
  \bibinfo{person}{Christoforos~I Mavrogiannis}, \bibinfo{person}{Julia Proft},
  {and} \bibinfo{person}{Claire Liang}.} \bibinfo{year}{2017}\natexlab{}.
\newblock \showarticletitle{Implicit communication in a joint action}. In
  \bibinfo{booktitle}{\emph{Proceedings of the 2017 acm/ieee international
  conference on human-robot interaction}}. ACM, \bibinfo{pages}{283--292}.
\newblock


\bibitem[\protect\citeauthoryear{Lafferty, Eady, and Elmers}{Lafferty
  et~al\mbox{.}}{1974}]%
        {lafferty1974desert}
\bibfield{author}{\bibinfo{person}{JC Lafferty}, \bibinfo{person}{Patrick~M
  Eady}, {and} \bibinfo{person}{J Elmers}.} \bibinfo{year}{1974}\natexlab{}.
\newblock \showarticletitle{The desert survival problem}.
\newblock \bibinfo{journal}{\emph{Experimental Learning Methods}}
  (\bibinfo{year}{1974}).
\newblock


\bibitem[\protect\citeauthoryear{Lasota, Fong, Shah, et~al\mbox{.}}{Lasota
  et~al\mbox{.}}{2017}]%
        {lasota2017survey}
\bibfield{author}{\bibinfo{person}{Przemyslaw~A Lasota},
  \bibinfo{person}{Terrence Fong}, \bibinfo{person}{Julie~A Shah},
  {et~al\mbox{.}}} \bibinfo{year}{2017}\natexlab{}.
\newblock \showarticletitle{A survey of methods for safe human-robot
  interaction}.
\newblock \bibinfo{journal}{\emph{Foundations and Trends{\textregistered} in
  Robotics}} \bibinfo{volume}{5}, \bibinfo{number}{4} (\bibinfo{year}{2017}),
  \bibinfo{pages}{261--349}.
\newblock


\bibitem[\protect\citeauthoryear{Lee}{Lee}{2018}]%
        {lee2018understanding}
\bibfield{author}{\bibinfo{person}{Min~Kyung Lee}.}
  \bibinfo{year}{2018}\natexlab{}.
\newblock \showarticletitle{Understanding perception of algorithmic decisions:
  Fairness, trust, and emotion in response to algorithmic management}.
\newblock \bibinfo{journal}{\emph{Big Data \& Society}} \bibinfo{volume}{5},
  \bibinfo{number}{1} (\bibinfo{year}{2018}),
  \bibinfo{pages}{2053951718756684}.
\newblock


\bibitem[\protect\citeauthoryear{Lee and Baykal}{Lee and Baykal}{2017}]%
        {lee2017algorithmic}
\bibfield{author}{\bibinfo{person}{Min~Kyung Lee} {and} \bibinfo{person}{Su
  Baykal}.} \bibinfo{year}{2017}\natexlab{}.
\newblock \showarticletitle{Algorithmic Mediation in Group Decisions: Fairness
  Perceptions of Algorithmically Mediated vs. Discussion-Based Social
  Division.}. In \bibinfo{booktitle}{\emph{CSCW}}. \bibinfo{pages}{1035--1048}.
\newblock


\bibitem[\protect\citeauthoryear{Lee, Kiesler, Forlizzi, and Rybski}{Lee
  et~al\mbox{.}}{2012}]%
        {lee2012ripple}
\bibfield{author}{\bibinfo{person}{Min~Kyung Lee}, \bibinfo{person}{Sara
  Kiesler}, \bibinfo{person}{Jodi Forlizzi}, {and} \bibinfo{person}{Paul
  Rybski}.} \bibinfo{year}{2012}\natexlab{}.
\newblock \showarticletitle{Ripple effects of an embedded social agent: a field
  study of a social robot in the workplace}. In
  \bibinfo{booktitle}{\emph{Proceedings of the SIGCHI Conference on Human
  Factors in Computing Systems}}. ACM, \bibinfo{pages}{695--704}.
\newblock


\bibitem[\protect\citeauthoryear{Moon, Troniak, Gleeson, Pan, Zheng, Blumer,
  MacLean, and Croft}{Moon et~al\mbox{.}}{2014}]%
        {moon2014meet}
\bibfield{author}{\bibinfo{person}{AJung Moon}, \bibinfo{person}{Daniel~M
  Troniak}, \bibinfo{person}{Brian Gleeson}, \bibinfo{person}{Matthew~KXJ Pan},
  \bibinfo{person}{Minhua Zheng}, \bibinfo{person}{Benjamin~A Blumer},
  \bibinfo{person}{Karon MacLean}, {and} \bibinfo{person}{Elizabeth~A Croft}.}
  \bibinfo{year}{2014}\natexlab{}.
\newblock \showarticletitle{Meet me where i'm gazing: how shared attention gaze
  affects human-robot handover timing}. In
  \bibinfo{booktitle}{\emph{Proceedings of the 2014 ACM/IEEE international
  conference on Human-robot interaction}}. ACM, \bibinfo{pages}{334--341}.
\newblock


\bibitem[\protect\citeauthoryear{Moreland}{Moreland}{2010}]%
        {moreland2010dyads}
\bibfield{author}{\bibinfo{person}{Richard~L Moreland}.}
  \bibinfo{year}{2010}\natexlab{}.
\newblock \showarticletitle{Are dyads really groups?}
\newblock \bibinfo{journal}{\emph{Small Group Research}} \bibinfo{volume}{41},
  \bibinfo{number}{2} (\bibinfo{year}{2010}), \bibinfo{pages}{251--267}.
\newblock


\bibitem[\protect\citeauthoryear{Morioka and Sakakibara}{Morioka and
  Sakakibara}{2010}]%
        {morioka2010new}
\bibfield{author}{\bibinfo{person}{Masashiro Morioka} {and}
  \bibinfo{person}{Shinsuke Sakakibara}.} \bibinfo{year}{2010}\natexlab{}.
\newblock \showarticletitle{A new cell production assembly system with
  human--robot cooperation}.
\newblock \bibinfo{journal}{\emph{CIRP annals}} \bibinfo{volume}{59},
  \bibinfo{number}{1} (\bibinfo{year}{2010}), \bibinfo{pages}{9--12}.
\newblock


\bibitem[\protect\citeauthoryear{Mutlu and Forlizzi}{Mutlu and
  Forlizzi}{2008}]%
        {mutlu2008robots}
\bibfield{author}{\bibinfo{person}{Bilge Mutlu} {and} \bibinfo{person}{Jodi
  Forlizzi}.} \bibinfo{year}{2008}\natexlab{}.
\newblock \showarticletitle{Robots in organizations: the role of workflow,
  social, and environmental factors in human-robot interaction}. In
  \bibinfo{booktitle}{\emph{Proceedings of the 3rd ACM/IEEE international
  conference on Human robot interaction}}. ACM, \bibinfo{pages}{287--294}.
\newblock


\bibitem[\protect\citeauthoryear{Mutlu, Shiwa, Kanda, Ishiguro, and
  Hagita}{Mutlu et~al\mbox{.}}{2009}]%
        {mutlu2009footing}
\bibfield{author}{\bibinfo{person}{Bilge Mutlu}, \bibinfo{person}{Toshiyuki
  Shiwa}, \bibinfo{person}{Takayuki Kanda}, \bibinfo{person}{Hiroshi Ishiguro},
  {and} \bibinfo{person}{Norihiro Hagita}.} \bibinfo{year}{2009}\natexlab{}.
\newblock \showarticletitle{Footing in human-robot conversations: how robots
  might shape participant roles using gaze cues}. In
  \bibinfo{booktitle}{\emph{Proceedings of the 4th ACM/IEEE international
  conference on Human robot interaction}}. ACM, \bibinfo{pages}{61--68}.
\newblock


\bibitem[\protect\citeauthoryear{Nikolaidis, Lasota, Rossano, Martinez,
  Fuhlbrigge, and Shah}{Nikolaidis et~al\mbox{.}}{2013}]%
        {nikolaidis2013human}
\bibfield{author}{\bibinfo{person}{Stefanos Nikolaidis},
  \bibinfo{person}{Przemyslaw Lasota}, \bibinfo{person}{Gregory Rossano},
  \bibinfo{person}{Carlos Martinez}, \bibinfo{person}{Thomas Fuhlbrigge}, {and}
  \bibinfo{person}{Julie Shah}.} \bibinfo{year}{2013}\natexlab{}.
\newblock \showarticletitle{Human-robot collaboration in manufacturing:
  Quantitative evaluation of predictable, convergent joint action}. In
  \bibinfo{booktitle}{\emph{Robotics (isr), 2013 44th international symposium
  on}}. IEEE, \bibinfo{pages}{1--6}.
\newblock


\bibitem[\protect\citeauthoryear{Oliveira, Arriaga, Alves-Oliveira, Correia,
  Petisca, and Paiva}{Oliveira et~al\mbox{.}}{2018}]%
        {oliveira2018friends}
\bibfield{author}{\bibinfo{person}{Raquel Oliveira},
  \bibinfo{person}{Patr{\'\i}cia Arriaga}, \bibinfo{person}{Patr{\'\i}cia
  Alves-Oliveira}, \bibinfo{person}{Filipa Correia}, \bibinfo{person}{Sofia
  Petisca}, {and} \bibinfo{person}{Ana Paiva}.}
  \bibinfo{year}{2018}\natexlab{}.
\newblock \showarticletitle{Friends or Foes?: Socioemotional Support and Gaze
  Behaviors in Mixed Groups of Humans and Robots}. In
  \bibinfo{booktitle}{\emph{Proceedings of the 2018 ACM/IEEE International
  Conference on Human-Robot Interaction}}. ACM, \bibinfo{pages}{279--288}.
\newblock


\bibitem[\protect\citeauthoryear{Reeves and Nass}{Reeves and Nass}{1996}]%
        {reeves1996media}
\bibfield{author}{\bibinfo{person}{Byron Reeves} {and}
  \bibinfo{person}{Clifford~Ivar Nass}.} \bibinfo{year}{1996}\natexlab{}.
\newblock \bibinfo{booktitle}{\emph{The media equation: How people treat
  computers, television, and new media like real people and places.}}
\newblock \bibinfo{publisher}{Cambridge university press}.
\newblock


\bibitem[\protect\citeauthoryear{Riek}{Riek}{2012}]%
        {riek2012wizard}
\bibfield{author}{\bibinfo{person}{Laurel~D Riek}.}
  \bibinfo{year}{2012}\natexlab{}.
\newblock \showarticletitle{Wizard of oz studies in hri: a systematic review
  and new reporting guidelines}.
\newblock \bibinfo{journal}{\emph{Journal of Human-Robot Interaction}}
  \bibinfo{volume}{1}, \bibinfo{number}{1} (\bibinfo{year}{2012}),
  \bibinfo{pages}{119--136}.
\newblock


\bibitem[\protect\citeauthoryear{Riether, Hegel, Wrede, and Horstmann}{Riether
  et~al\mbox{.}}{2012}]%
        {riether2012social}
\bibfield{author}{\bibinfo{person}{Nina Riether}, \bibinfo{person}{Frank
  Hegel}, \bibinfo{person}{Britta Wrede}, {and} \bibinfo{person}{Gernot
  Horstmann}.} \bibinfo{year}{2012}\natexlab{}.
\newblock \showarticletitle{Social facilitation with social robots?}. In
  \bibinfo{booktitle}{\emph{Proceedings of the seventh annual ACM/IEEE
  international conference on Human-Robot Interaction}}. ACM,
  \bibinfo{pages}{41--48}.
\newblock


\bibitem[\protect\citeauthoryear{Saupp{\'e} and Mutlu}{Saupp{\'e} and
  Mutlu}{2015}]%
        {sauppe2015social}
\bibfield{author}{\bibinfo{person}{Allison Saupp{\'e}} {and}
  \bibinfo{person}{Bilge Mutlu}.} \bibinfo{year}{2015}\natexlab{}.
\newblock \showarticletitle{The social impact of a robot co-worker in
  industrial settings}. In \bibinfo{booktitle}{\emph{Proceedings of the 33rd
  annual ACM conference on human factors in computing systems}}. ACM,
  \bibinfo{pages}{3613--3622}.
\newblock


\bibitem[\protect\citeauthoryear{Scassellati, Admoni, and
  Matari{\'c}}{Scassellati et~al\mbox{.}}{2012}]%
        {scassellati2012robots}
\bibfield{author}{\bibinfo{person}{Brian Scassellati}, \bibinfo{person}{Henny
  Admoni}, {and} \bibinfo{person}{Maja Matari{\'c}}.}
  \bibinfo{year}{2012}\natexlab{}.
\newblock \showarticletitle{Robots for use in autism research}.
\newblock \bibinfo{journal}{\emph{Annual review of biomedical engineering}}
  \bibinfo{volume}{14} (\bibinfo{year}{2012}), \bibinfo{pages}{275--294}.
\newblock


\bibitem[\protect\citeauthoryear{Scheinman}{Scheinman}{1969}]%
        {scheinman1969design}
\bibfield{author}{\bibinfo{person}{Victor~David Scheinman}.}
  \bibinfo{year}{1969}\natexlab{}.
\newblock \bibinfo{booktitle}{\emph{Design of a computer controlled
  manipulator.}}
\newblock \bibinfo{type}{{T}echnical {R}eport}. \bibinfo{institution}{Stanford
  Univ Calif Dept of Computer Science}.
\newblock


\bibitem[\protect\citeauthoryear{Scheutz, Schermerhorn, and Kramer}{Scheutz
  et~al\mbox{.}}{2006}]%
        {scheutz2006utility}
\bibfield{author}{\bibinfo{person}{Matthias Scheutz}, \bibinfo{person}{Paul
  Schermerhorn}, {and} \bibinfo{person}{James Kramer}.}
  \bibinfo{year}{2006}\natexlab{}.
\newblock \showarticletitle{The utility of affect expression in natural
  language interactions in joint human-robot tasks}. In
  \bibinfo{booktitle}{\emph{Proceedings of the 1st ACM SIGCHI/SIGART conference
  on Human-robot interaction}}. ACM, \bibinfo{pages}{226--233}.
\newblock


\bibitem[\protect\citeauthoryear{Shah, Wiken, Williams, and Breazeal}{Shah
  et~al\mbox{.}}{2011}]%
        {shah2011improved}
\bibfield{author}{\bibinfo{person}{Julie Shah}, \bibinfo{person}{James Wiken},
  \bibinfo{person}{Brian Williams}, {and} \bibinfo{person}{Cynthia Breazeal}.}
  \bibinfo{year}{2011}\natexlab{}.
\newblock \showarticletitle{Improved human-robot team performance using chaski,
  a human-inspired plan execution system}. In
  \bibinfo{booktitle}{\emph{Proceedings of the 6th international conference on
  Human-robot interaction}}. ACM, \bibinfo{pages}{29--36}.
\newblock


\bibitem[\protect\citeauthoryear{Shah}{Shah}{2017}]%
        {shah2017enhancing}
\bibfield{author}{\bibinfo{person}{Julie~A Shah}.}
  \bibinfo{year}{2017}\natexlab{}.
\newblock \showarticletitle{Enhancing Human Capability with Intelligent Machine
  Teammates}. In \bibinfo{booktitle}{\emph{Proceedings of the 16th Conference
  on Autonomous Agents and MultiAgent Systems}}. International Foundation for
  Autonomous Agents and Multiagent Systems, \bibinfo{pages}{4--4}.
\newblock


\bibitem[\protect\citeauthoryear{Shen, Slovak, and Jung}{Shen
  et~al\mbox{.}}{2018}]%
        {shen2018stop}
\bibfield{author}{\bibinfo{person}{Solace Shen}, \bibinfo{person}{Petr Slovak},
  {and} \bibinfo{person}{Malte~F Jung}.} \bibinfo{year}{2018}\natexlab{}.
\newblock \showarticletitle{Stop. I See a Conflict Happening.: A Robot Mediator
  for Young Children's Interpersonal Conflict Resolution}. In
  \bibinfo{booktitle}{\emph{Proceedings of the 2018 ACM/IEEE International
  Conference on Human-Robot Interaction}}. ACM, \bibinfo{pages}{69--77}.
\newblock


\bibitem[\protect\citeauthoryear{Short, Swift-Spong, Shim, Wisniewski, Zak, Wu,
  Zelinski, and Matari{\'c}}{Short et~al\mbox{.}}{2017}]%
        {short2017understanding}
\bibfield{author}{\bibinfo{person}{Elaine~Schaertl Short},
  \bibinfo{person}{Katelyn Swift-Spong}, \bibinfo{person}{Hyunju Shim},
  \bibinfo{person}{Kristi~M Wisniewski}, \bibinfo{person}{Deanah~Kim Zak},
  \bibinfo{person}{Shinyi Wu}, \bibinfo{person}{Elizabeth Zelinski}, {and}
  \bibinfo{person}{Maja~J Matari{\'c}}.} \bibinfo{year}{2017}\natexlab{}.
\newblock \showarticletitle{Understanding social interactions with socially
  assistive robotics in intergenerational family groups}. In
  \bibinfo{booktitle}{\emph{Robot and Human Interactive Communication (RO-MAN),
  2017 26th IEEE International Symposium on}}. IEEE, \bibinfo{pages}{236--241}.
\newblock


\bibitem[\protect\citeauthoryear{Siino and Hinds}{Siino and Hinds}{2005}]%
        {siino2005robots}
\bibfield{author}{\bibinfo{person}{Rosanne~M Siino} {and}
  \bibinfo{person}{Pamela~J Hinds}.} \bibinfo{year}{2005}\natexlab{}.
\newblock \showarticletitle{Robots, gender \& sensemaking: Sex segregation’s
  impact on workers making sense of a mobile autonomous robot}. In
  \bibinfo{booktitle}{\emph{Robotics and Automation, 2005. ICRA 2005.
  Proceedings of the 2005 IEEE International Conference on}}. IEEE,
  \bibinfo{pages}{2773--2778}.
\newblock


\bibitem[\protect\citeauthoryear{Simmel}{Simmel}{1964}]%
        {simmel1964sociology}
\bibfield{author}{\bibinfo{person}{Georg Simmel}.}
  \bibinfo{year}{1964}\natexlab{}.
\newblock \bibinfo{booktitle}{\emph{The sociology of georg simmel}}.
  Vol.~\bibinfo{volume}{92892}.
\newblock \bibinfo{publisher}{Simon and Schuster}.
\newblock


\bibitem[\protect\citeauthoryear{Steinfeld, Fong, Kaber, Lewis, Scholtz,
  Schultz, and Goodrich}{Steinfeld et~al\mbox{.}}{2006}]%
        {steinfeld2006common}
\bibfield{author}{\bibinfo{person}{Aaron Steinfeld}, \bibinfo{person}{Terrence
  Fong}, \bibinfo{person}{David Kaber}, \bibinfo{person}{Michael Lewis},
  \bibinfo{person}{Jean Scholtz}, \bibinfo{person}{Alan Schultz}, {and}
  \bibinfo{person}{Michael Goodrich}.} \bibinfo{year}{2006}\natexlab{}.
\newblock \showarticletitle{Common metrics for human-robot interaction}. In
  \bibinfo{booktitle}{\emph{Proceedings of the 1st ACM SIGCHI/SIGART conference
  on Human-robot interaction}}. ACM, \bibinfo{pages}{33--40}.
\newblock


\bibitem[\protect\citeauthoryear{Strohkorb~Sebo, Traeger, Jung, and
  Scassellati}{Strohkorb~Sebo et~al\mbox{.}}{2018}]%
        {strohkorb2018ripple}
\bibfield{author}{\bibinfo{person}{Sarah Strohkorb~Sebo},
  \bibinfo{person}{Margaret Traeger}, \bibinfo{person}{Malte Jung}, {and}
  \bibinfo{person}{Brian Scassellati}.} \bibinfo{year}{2018}\natexlab{}.
\newblock \showarticletitle{The Ripple Effects of Vulnerability: The Effects of
  a Robot's Vulnerable Behavior on Trust in Human-Robot Teams}. In
  \bibinfo{booktitle}{\emph{Proceedings of the 2018 ACM/IEEE International
  Conference on Human-Robot Interaction}}. ACM, \bibinfo{pages}{178--186}.
\newblock


\bibitem[\protect\citeauthoryear{Stubbs, Hinds, and Wettergreen}{Stubbs
  et~al\mbox{.}}{2007}]%
        {stubbs2007autonomy}
\bibfield{author}{\bibinfo{person}{Kristen Stubbs}, \bibinfo{person}{Pamela~J
  Hinds}, {and} \bibinfo{person}{David Wettergreen}.}
  \bibinfo{year}{2007}\natexlab{}.
\newblock \showarticletitle{Autonomy and common ground in human-robot
  interaction: A field study}.
\newblock \bibinfo{journal}{\emph{IEEE Intelligent Systems}}
  \bibinfo{volume}{22}, \bibinfo{number}{2} (\bibinfo{year}{2007}).
\newblock


\bibitem[\protect\citeauthoryear{Takayama, Dooley, and Ju}{Takayama
  et~al\mbox{.}}{2011}]%
        {takayama2011expressing}
\bibfield{author}{\bibinfo{person}{Leila Takayama}, \bibinfo{person}{Doug
  Dooley}, {and} \bibinfo{person}{Wendy Ju}.} \bibinfo{year}{2011}\natexlab{}.
\newblock \showarticletitle{Expressing thought: improving robot readability
  with animation principles}. In \bibinfo{booktitle}{\emph{Human-Robot
  Interaction (HRI), 2011 6th ACM/IEEE International Conference on}}. IEEE,
  \bibinfo{pages}{69--76}.
\newblock


\bibitem[\protect\citeauthoryear{V{\'a}zquez, Carter, McDorman, Forlizzi,
  Steinfeld, and Hudson}{V{\'a}zquez et~al\mbox{.}}{2017}]%
        {vazquez2017towards}
\bibfield{author}{\bibinfo{person}{Marynel V{\'a}zquez},
  \bibinfo{person}{Elizabeth~J Carter}, \bibinfo{person}{Braden McDorman},
  \bibinfo{person}{Jodi Forlizzi}, \bibinfo{person}{Aaron Steinfeld}, {and}
  \bibinfo{person}{Scott~E Hudson}.} \bibinfo{year}{2017}\natexlab{}.
\newblock \showarticletitle{Towards robot autonomy in group conversations:
  Understanding the effects of body orientation and gaze}. In
  \bibinfo{booktitle}{\emph{Proceedings of the 2017 ACM/IEEE International
  Conference on Human-Robot Interaction}}. ACM, \bibinfo{pages}{42--52}.
\newblock


\bibitem[\protect\citeauthoryear{Vertesi}{Vertesi}{2015}]%
        {vertesi2015seeing}
\bibfield{author}{\bibinfo{person}{Janet Vertesi}.}
  \bibinfo{year}{2015}\natexlab{}.
\newblock \bibinfo{booktitle}{\emph{Seeing like a rover: How robots, teams, and
  images craft knowledge of Mars}}.
\newblock \bibinfo{publisher}{University of Chicago Press}.
\newblock


\bibitem[\protect\citeauthoryear{Wilcox, Nikolaidis, and Shah}{Wilcox
  et~al\mbox{.}}{2013}]%
        {wilcox2013optimization}
\bibfield{author}{\bibinfo{person}{Ronald Wilcox}, \bibinfo{person}{Stefanos
  Nikolaidis}, {and} \bibinfo{person}{Julie Shah}.}
  \bibinfo{year}{2013}\natexlab{}.
\newblock \showarticletitle{Optimization of temporal dynamics for adaptive
  human-robot interaction in assembly manufacturing}.
\newblock \bibinfo{journal}{\emph{Robotics}} (\bibinfo{year}{2013}),
  \bibinfo{pages}{441}.
\newblock


\bibitem[\protect\citeauthoryear{Williams}{Williams}{1997}]%
        {williams1997social}
\bibfield{author}{\bibinfo{person}{Kipling~D Williams}.}
  \bibinfo{year}{1997}\natexlab{}.
\newblock \showarticletitle{Social ostracism}.
\newblock In \bibinfo{booktitle}{\emph{Aversive interpersonal behaviors}}.
  \bibinfo{publisher}{Springer}, \bibinfo{pages}{133--170}.
\newblock


\bibitem[\protect\citeauthoryear{Williams}{Williams}{2010}]%
        {williams2010dyads}
\bibfield{author}{\bibinfo{person}{Kipling~D Williams}.}
  \bibinfo{year}{2010}\natexlab{}.
\newblock \showarticletitle{Dyads can be groups (and often are)}.
\newblock \bibinfo{journal}{\emph{Small Group Research}} \bibinfo{volume}{41},
  \bibinfo{number}{2} (\bibinfo{year}{2010}), \bibinfo{pages}{268--274}.
\newblock


\bibitem[\protect\citeauthoryear{Wobbrock}{Wobbrock}{2012}]%
        {wobbrock2012seven}
\bibfield{author}{\bibinfo{person}{Jacob~O Wobbrock}.}
  \bibinfo{year}{2012}\natexlab{}.
\newblock \showarticletitle{Seven research contributions in HCI}.
\newblock \bibinfo{journal}{\emph{studies}} \bibinfo{volume}{1},
  \bibinfo{number}{1} (\bibinfo{year}{2012}), \bibinfo{pages}{52--80}.
\newblock


\bibitem[\protect\citeauthoryear{Wobbrock and Kientz}{Wobbrock and
  Kientz}{2016}]%
        {wobbrock2016research}
\bibfield{author}{\bibinfo{person}{Jacob~O Wobbrock} {and}
  \bibinfo{person}{Julie~A Kientz}.} \bibinfo{year}{2016}\natexlab{}.
\newblock \showarticletitle{Research contributions in human-computer
  interaction}.
\newblock \bibinfo{journal}{\emph{interactions}} \bibinfo{volume}{23},
  \bibinfo{number}{3} (\bibinfo{year}{2016}), \bibinfo{pages}{38--44}.
\newblock


\bibitem[\protect\citeauthoryear{Woods, Walters, Koay, and Dautenhahn}{Woods
  et~al\mbox{.}}{2006}]%
        {woods2006methodological}
\bibfield{author}{\bibinfo{person}{S Woods}, \bibinfo{person}{Michael~L
  Walters}, \bibinfo{person}{Kheng~Lee Koay}, {and} \bibinfo{person}{Kerstin
  Dautenhahn}.} \bibinfo{year}{2006}\natexlab{}.
\newblock \showarticletitle{Methodological issues in HRI: A comparison of live
  and video-based methods in robot to human approach direction trials}. In
  \bibinfo{booktitle}{\emph{Procs 15th IEEE Int Symp on Robot and Human
  Interactive Communication, ROMAN}}.
\newblock


\bibitem[\protect\citeauthoryear{You and Robert}{You and Robert}{2017}]%
        {you2017emotional}
\bibfield{author}{\bibinfo{person}{Sangseok You} {and} \bibinfo{person}{Lionel
  Robert}.} \bibinfo{year}{2017}\natexlab{}.
\newblock \showarticletitle{Emotional attachment, performance, and viability in
  teams collaborating with embodied physical action (EPA) robots}. AIS.
\newblock


\end{thebibliography}

\end{document}